\documentclass[lettersize,journal]{IEEEtran}
\usepackage{amsmath,amsfonts}
\usepackage{array}
\usepackage[caption=false,font=normalsize,labelfont=sf,textfont=sf]{subfig}
\usepackage{textcomp}
\usepackage{stfloats}
\usepackage{url}
\usepackage{verbatim}
\usepackage{graphicx}

\usepackage{multirow}
\usepackage{marvosym}
\usepackage{algorithm}
\usepackage{algpseudocode}
\usepackage{xcolor}
\usepackage{courier}
\usepackage{indentfirst}
\usepackage{amsmath}
\usepackage{amssymb}
\usepackage{booktabs}
\usepackage{balance}

\def\BibTeX{{\rm B\kern-.05em{\sc i\kern-.025em b}\kern-.08em
    T\kern-.1667em\lower.7ex\hbox{E}\kern-.125emX}}

\begin{document}
\title{G$^2$V$^2$former: Graph Guided Video Vision Transformer for Face Anti-Spoofing}

\author{
    Jingyi Yang, Zitong Yu,~\IEEEmembership{Senior Member,~IEEE}, Jia He, Xiuming Ni, Liepiao Zhang, Hui Li, \\Xiaochun Cao,~\IEEEmembership{Senior Member,~IEEE}
\IEEEcompsocitemizethanks{
\IEEEcompsocthanksitem Received December 2024. This work was supported by the National Science Foundation of China, under Grant No. 62171425, and Guangdong Basic and Applied Basic Research Foundation (Grant No. 2023A1515140037). Corresponding Authors: Zitong Yu and Hui Li. 
\IEEEcompsocthanksitem Jingyi Yang and Hui Li are with University of Science and Technology of China, Hefei, China. (email: yangjingyi@mail.ustc.edu.cn and mythlee@ustc.edu.cn)
\IEEEcompsocthanksitem Zitong Yu is with Great Bay University, Dongguan, China. (email: zitong.yu@ieee.org)
\IEEEcompsocthanksitem Jia He and Xiuming Ni are with Anhui Tsinglink Information Tech. Com Ltd., Hefei, China. (email: hejia@ustc.edu and nixm@tsinglink.com)
\IEEEcompsocthanksitem Liepiao Zhang is with South China University of Technology, Guangzhou, China. (email: auzlpiao@mail.scut.edu.cn)
\IEEEcompsocthanksitem Xiaochun Cao is with Shenzhen Campus of Sun Yat-sen University, Shenzhen, China. (email: caoxiaochun@mail.sysu.edu.cn)
}}



\maketitle

\begin{abstract}
In videos containing spoofed faces, we may uncover the spoofing evidence based on either photometric or dynamic abnormality, or a combination of both. Prevailing face anti-spoofing (FAS) approaches generally concentrate on the single-frame scenario; however, purely photometric-driven methods overlook the dynamic spoofing clues that may be exposed over time. This may lead FAS systems to conclude incorrect judgments, especially in cases where it is easily distinguishable in terms of dynamics but challenging to discern in terms of photometrics. To this end, we propose the Graph Guided Video Vision Transformer (G\(^2\)V\(^2\)former), which combines faces with facial landmarks for the fusion of photometric and dynamic features. We factorize the attention into space and time and fuse them via a spatiotemporal block. Specifically, we design a novel temporal attention, called Kronecker temporal attention, which has a wider receptive field and is beneficial for capturing dynamic information. Moreover, we leverage the low-semantic motion of facial landmarks to guide the high-semantic change of facial expressions based on the motivation that regions containing landmarks may reveal more dynamic clues. Extensive experiments on nine benchmark datasets demonstrate that our method achieves superior performance under various scenarios. The codes will be released soon.
\end{abstract}

\section{Introduction}
\label{sec:intro}
\IEEEPARstart{F}{ace} recognition (FR), as the most notable and successful biometric recognition technology, has been widely applied in different scenarios such as access control and mobile electronic payments. Despite its success, FR systems are vulnerable to various presentation attacks, including printed photos, video replay, 3D masks, etc. Fortunately, face anti-spoofing (FAS) techniques have been proven to play a crucial role in enhancing the security of FR systems. Consequently, this topic has gradually emerged as a recent research hotspot.

\begin{figure}[t]
    \centering
    \includegraphics[width=0.49\textwidth,height=0.16\textwidth]{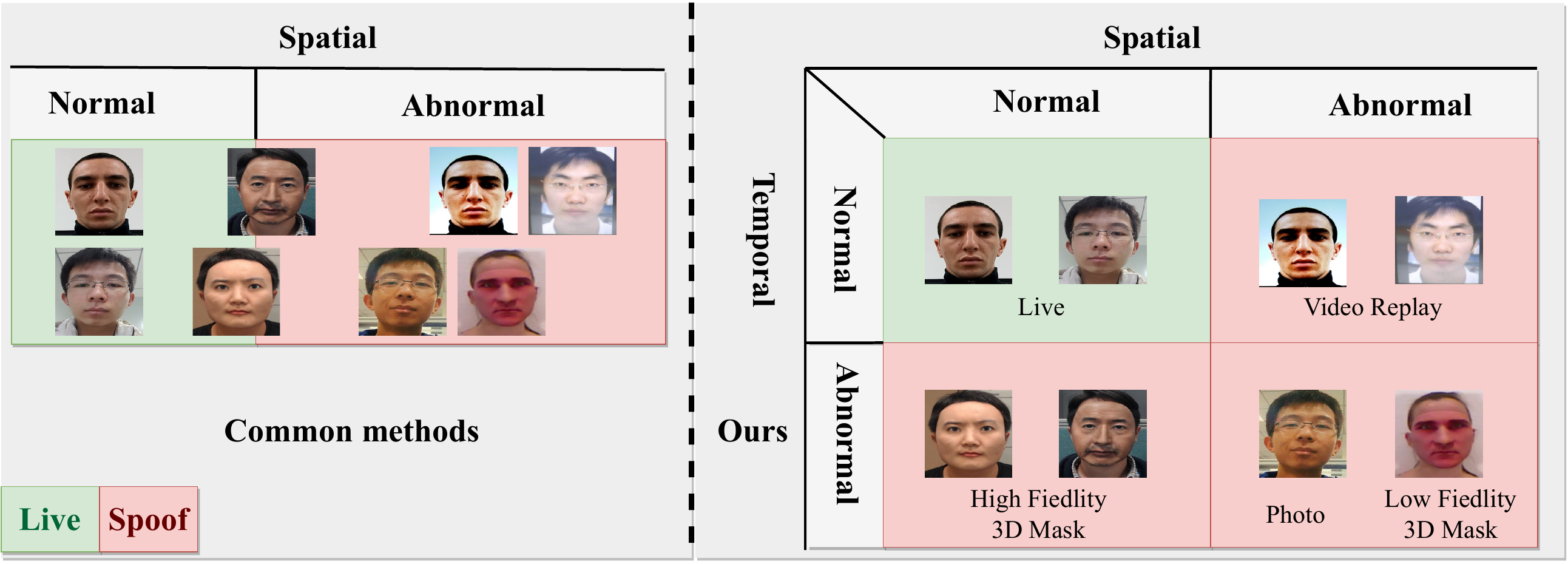}
\vspace{-0.5cm}
    \caption{Most common methods in FAS foucs on frame-level spoofing representation. In contrast, our method aim to fuse photometric and dynamic spoofing clues. When the testing sample shows abnormalities in any dimension (space or time), it should be detected as a spoof.}
    \label{fig: Motivation}
\vspace{-0.3cm}
\end{figure}

Most previous research focuses on frame-level anti-spoofing, including but not limited to extracting hand-crafted texture \cite{boulkenafet2015face,patel2016secure,yang2013face}, and deeply learned representations \cite{li2016original,yu2020searching}, adopting domain generalization \cite{shao2019multi,jia2020single,chen2021generalizable,liu2021adaptive,wang2022domain,zhou2023instance,sun2023rethinking,yang2024generalized}, and domain adaptation \cite{wang2020unsupervised,wang2021self} techniques, attempting meta-learning paradigms \cite{shao2020regularized,wang2021self,liu2021adaptive}, and so on. Generally speaking, their common goal is to capture the photometric distinction between live and spoof faces, hoping that it is universal. Despite their significant success in FAS tasks, they overlook the possibility that spoof clues may be exposed over time. A pure photometric-driven algorithm has not fully exploited the spoofing information contained in videos and might be insufficient to represent the intrinsic gap between the live and spoof samples, as shown to the left of Fig. \ref{fig: Motivation}. These methods may be vulnerable to spoof attacks that are difficult to distinguish in terms of photometric, such as high fidelity 3D-masks.

There are already several efforts to explore spatiotemporal anti-spoofing feature at the video level. Some research exploits the models composed of CNN-LSTM/GRU \cite{liu2018learning,saha2020domain,wang2020deep,yang2019face}, 3D-CNNs \cite{wang2021multi,xu2021exploiting}, or transformers \cite{ming2022vitranspad,khan2021video,wang2022learning,khan2021video,wang2022learning}. Recently, GAIN \cite{chang2023closer} proposes combining the skeleton-based action recognition with photometric-based face anti-spoofing. It utilizes ST-GCN \cite{yan2018spatial} to capture the motion information contained in facial landmarks and integrates it with existing photometric-based FAS frameworks, showing an extraordinary performance. However, landmarks are only three-dimensional coordinates (low-semantic), may not be enough to carry complex dynamic information like pixel motion.

\begin{figure*}[t]
    \centering
    \includegraphics[width=0.9\textwidth,height=0.37\textwidth]{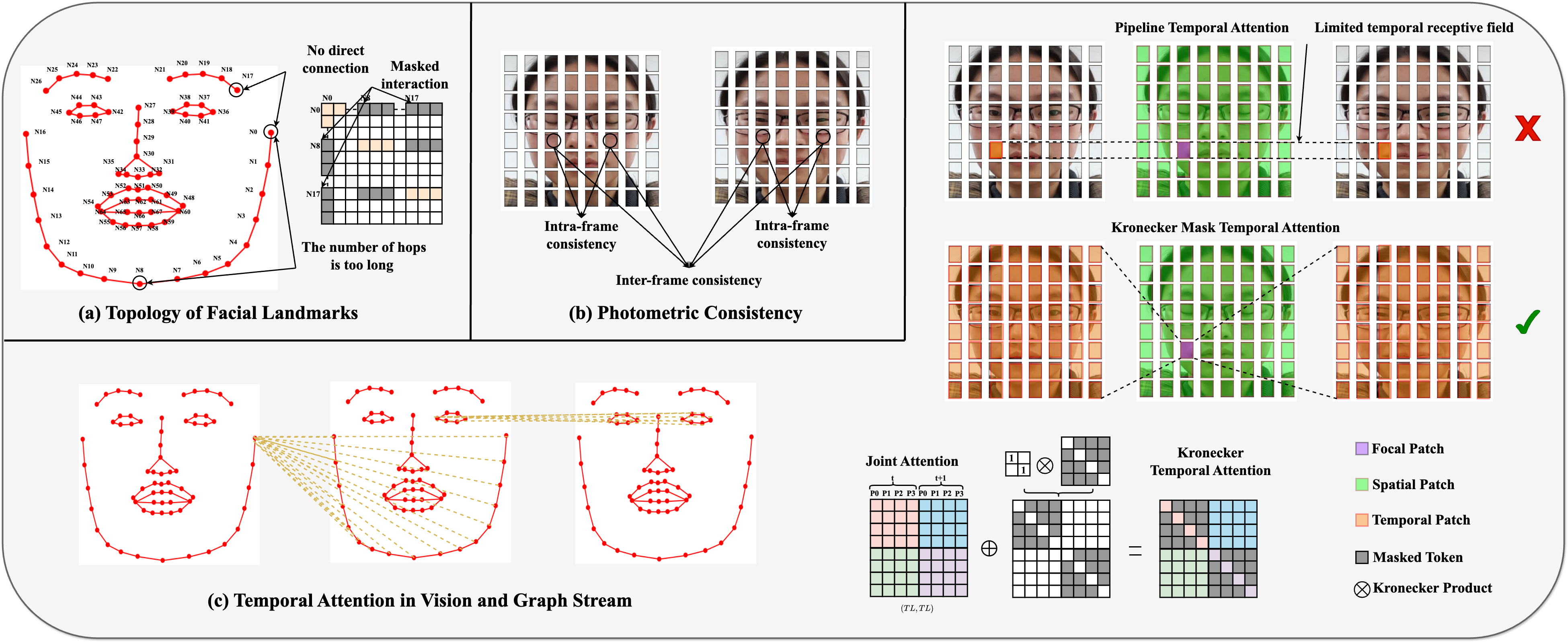}
    \caption{(a) The topology-aware spatial attention. Connections between two nodes will be blocked (masked in attention matrix) if they are too far apart in the topology. (b) Various visual spatiotemporal attention. (c) Common spatial attention can be seen as applying a mask to joint attention. (d) Kronecker temporal attention can also be obtained by employing a tailored mask to joint attention. Spatial attention and Kronecker temporal attention exhibit spatiotemporal complementarity.}
    \label{fig: Spatial Temporal Forms}
\end{figure*}

Aiming to fully exploit spoofing clues in both spatial and temporal dimensions, we propose a transformer-based architecture that takes both images and landmarks as inputs to integrate photometric feature with dynamic information. We factorize the attention to spatial and temporal dimensions, where the spatial attention in vision aggregates the photometric feature and topology-aware spatial attention serves for the perception of facial contours through landmarks, as shown in Fig. \ref{fig: Spatial Temporal Forms} (a) and (b). Our proposed Kronecker temporal attention exhibits a wider receptive field, and is conducive for global dynamic information capture, as illustrated in Fig. \ref{fig: Spatial Temporal Forms} (c). We extract rough facial motion through temporal attention of landmarks and then guide the temporal attention of vision branch to focus on patches that are more likely to occur pixel motion. Our main contributions are three-folds:
\begin{itemize}
\item We propose a novel architecture Graph Guided Video Vision Transformer (G\(^2\)V\(^2\)former) to fuse the spatial and temporal features from video and landmarks for face anti-spoofing tasks as depicted to the right of Fig. \ref{fig: Motivation}, achieving remarkable generalization.
\item We design a novel Kronecker temporal attention which is more conducive to capturing the patch or node relationship of different timestamps and is beneficial to capturing comprehensive temporal clues.
\item We propose the graph-guided vision temporal attention which leverages temporal attention of landmarks motion to guide the capture of the higher-semantic pixel motion.
\end{itemize}

\section{Related Work}
\label{sec:related work}
In this section, we first provide some detailed discussion about the face anti-spoofing, especially those focusing on temporal clue mining. After that, we give a brief review of the video vision transformer family. Finally, we discuss the ST-GCN and Graphormer.

\subsection{Mining Temporal Clues for Face Anti-Spoofing}
Numerous face anti-spoofing methods \cite{li2016generalized,liu2018learning,liu2018remote,yu2021transrppg} attempt to exploit temporal information within video clips, particularly through the analysis of representative biological signals, such as Remote Photoplethysmography (rPPG) signal. Since rPPG signals are detectable only in live individuals, they provide a means of distinguishing genuine faces from spoofed ones. However, these approaches heavily depend on the credibility and robustness of the extracted rPPG signals, requiring a sufficiently long video. Other methods leverage changes in facial expressions, including eyes blinking \cite{patel2016cross} and lip or mouth movements \cite{kollreider2007real,shao2017deep}. Additionally, a range of approaches \cite{liu2018learning,saha2020domain,wang2020deep,yang2019face,wang2021multi,xu2021exploiting} propose to capture temporal correlation by designing specialized network architectures. While recent studies \cite{ming2022vitranspad,khan2021video,wang2022learning} incorporate transformer-based architectures for FAS task. For instance, Khan et al. \cite{khan2021video} introduce a video transformer with incremental learning for detecting deepfake videos, and TTN \cite{wang2022learning} employs a temporal transformer network and a temporal depth difference loss to learn multi-granularity temporal characteristics for FAS. More recently, GAIN \cite{chang2023closer} utilizes ST-GCN \cite{yan2018spatial} to extract the deeply learned spatial-temporal features from facial landmarks, alongside photometric features derived from existing photometric-based methods. In this work, we aim to effectively extract robust spatiotemporal representation from both spatial dimension and temporal dimension, and spatiotemporal graph representation.

\subsection{Video Vision Transformer}
The self-attention mechanism, known for its excellent ability to model long-range dependencies, provides greater flexibility and effectiveness compared to 3D convolution when processing videos and performing spatiotemporal fusion. Efforts to extend Vision Transformer (ViT) \cite{dosovitskiy2020image} to the video domain include approaches such as \cite{bertasius2021space,arnab2021vivit,neimark2021video,guo2021ssan,liu2022video,yang2025kronecker}, which propose various forms of spatiotemporal attention. These can be broadly categorized into two types: joint and divided (spatio and temporal) attention. Furthermore, the pioneering work MAE \cite{he2022masked} demonstrated that applying masked content prediction for self-supervised learning in visual tasks achieves significant benefits. Similar to how ViT extends naturally to the video domain, MAE has been adapted for video tasks \cite{feichtenhofer2022masked,tong2022videomae}. These methods typically employ a higher masking ratio due to the redundancy inherent in videos. Compared to images, video data offers more masking strategies, and \cite{feichtenhofer2022masked} explores four common masking strategies. In this article, we propose a new temporal modeling approach for visual temporal attention and spatiotemporal graph temporal attention.

\subsection{ST-GCN and Graphormer}
Spatiotemporal graphs are widely utilized in skeleton-based action recognition to model dynamic information derived from joint trajectories. ST-GCN \cite{yan2018spatial} employs Graph Convolutional Networks (GCNs) \cite{kipf2016semi} in combination with temporal convolutions to extract robust dynamic representation from joint sequences. Various studies \cite{yan2018spatial,chen2021channel,liu2020disentangling,shi2019two} have proven the strong modeling capabilities of GCNs for topological relationships. Recently, Graphormer \cite{ying2021transformers} leveraged the self-attention mechanisms in modeling graph topology relationships, achieving remarkable performance. It demonstrates that many Graph Neural Networks (GNNs) can be seen as special cases of Graphormer. Specifically, the normalized adjacency matrix in GNNs can be regarded as a special self-attention matrix, the functionality of the normalized adjacency matrix can also be substituted by the self-attention matrix with topological knowledge, and the self-attention matrix exhibits stronger expressive power. In this paper, we extend Graphormer \cite{ying2021transformers} to 'ST-Graphormer' for spatiotemporal graph representation fusion.

\begin{figure*}[t!]
    \centering
    \includegraphics[width=0.97\textwidth,height=0.4\textwidth]{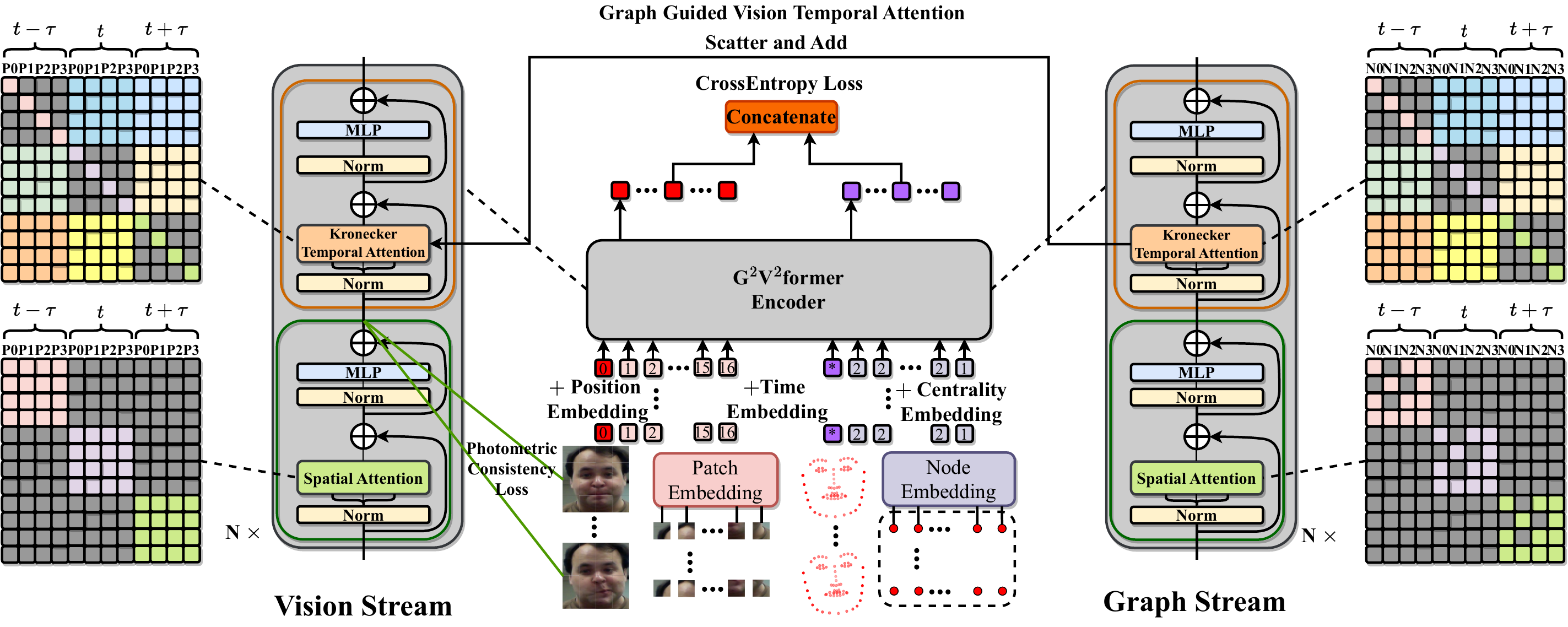}
    \vspace{-0.3cm}
    \caption{The architecture of graph guided video vision transformer. It requires inputting a facial video clip into the vision stream and the corresponding facial landmarks into the graph stream. The visual spatial attention is guided by photometric consistency loss, while the visual temporal attention is guided by graphic temporal attention via scatter and add operation. We concatenate the head token of both stream for classification, and optimized it through cross-entropy loss.}
    \label{fig: Architecture}
\vspace{-0.2cm}
\end{figure*}

\section{Methodology}
\label{sec: methodology}
\noindent The Graph Guided Video Vision Transformer (G$^2$V$^2$former) is a two-stream architecture designed to fuse both static and dynamic spoofing clues derived from facial appearances and landmarks. The model employs a cascaded spatiotemporal attention form, as shown in Fig. \ref{fig: Architecture}. Each spatiotemporal block (STB) in the vision stream or graph stream consists of a spatial attention module and a temporal attention module, enabling effective spatiotemporal feature fusion. Both the vision stream and graph stream are constructed by stacking multiple STBs sequentially. The training process of G$^2$V$^2$former is divided into two stages. In the first stage, we follow MAE's pre-training paradigm, conducting independent pre-training for the vision and graph streams. During the second stage, we fine-tune the G$^2$V$^2$former, incorporating the graph stream as a guide to enhance the learning of the vision stream.

During pre-training, a high masking ratio of 90\% is applied independently to both the vision and graph streams. Specifically, this ratio is maintained for each frame, ensuring uniform masking across frames, though the masked positions may vary. This strategy ensures comprehensive training of both the spatial and temporal attention modules. Additionally, absolute positional and temporal embeddings are incorporated, as illustrated in Fig. \ref{fig: Architecture}. In the vision stream, we add absolute position embedding to the patches at the corresponding positions of each frame and add absolute time embedding along the timeline to the patches of different frames. Specifically, patches from the same frame sharing an unique absolute time embedding, while different timestamps have different embedding values:
\begin{equation}
\label{eqn: position time embedding}
\hat{\textbf{v}}^t_i = \textbf{v}^t_i + \textbf{e}^v_{\mathrm{pos}(i)} + \textbf{e}^v_{\mathrm{tim}(t)},
\end{equation}
where \(\textbf{v}\) represents embedded patch, \(\hat{\textbf{v}}\) is the output after positional embedding, superscript \(t\) denotes the \(t\)-th timestamp. Subscript \(i\) denotes the patch sequence number. \(\textbf{e}^v_{\mathrm{pos}(i)}\) is the absolute position embedding of \(i\)-th patches, \(\textbf{e}^v_{\mathrm{tim}(t)}\) is the absolute time embedding of patches belong to \(t\)-th frame.

In the graph stream, centrality embedding and absolute time embedding are involved:
\begin{equation}
\label{eqn: position time embedding of graph}
\hat{\textbf{g}}^t_i = \textbf{g}^t_i + \textbf{e}^g_{\mathrm{cen}(i)} + \textbf{e}^g_{\mathrm{tim}(t)},
\end{equation}
where \(\textbf{g}\) represents input embedding, \(\hat{\textbf{g}}\) is output embedding. \(\textbf{e}^g_{\mathrm{tim}(t)}\) is the absolute time embedding of nodes belong to \(t\)-th frame, \(\textbf{e}_{\mathrm{cen}(i)}\) is the centrality embedding which corresponds to a lookup table, that is, when the degree of the current node is $i$, the corresponding centrality embedding will be output based on the index $i$ of the lookup table. Its function is to help the model understand the importance between nodes. We will later mention this part at Sec. \ref{subsec:spatial attention}.

To better capture high-frequency variations in the three-dimensional (x-y-z) coordinates of facial landmarks, we adopt the sine-cosine positional encoding derived from \cite{mildenhall2021nerf}, where \(F\) is the number of frequencies:
\begin{equation}
\label{eqn: sin cos positional encoding}
\small
\gamma(x) = (\mathrm{sin}(2^0 \pi x),\mathrm{cos}(2^0 \pi x), \cdots, \mathrm{sin}(2^{F-1} \pi x),\mathrm{cos}(2^{F-1} \pi x))
\end{equation}

In the fine-tuning stage, we concatenate the class tokens of vision stream with the hyper-node tokens from the graph stream. The combined tokens are then utilized for classification.

\subsection{Photometric Consistency in Spatial Attention}
\label{subsec:spatial attention}
First, we implement patch embedding \cite{dosovitskiy2020image}, where each image is divided into non-overlapping patches, flattened, and passed through standard self-attention calculations. 
Spatial Attention (SA) can be formulated as:
\begin{equation}
\label{eqn: spatial attention 0}
\textbf{Q}_s=\textbf{X}_{(l \times d)}\textbf{W}_{\mathrm{Q}}, 
\textbf{K}_s=\textbf{X}_{(l \times d)}\textbf{W}_{\mathrm{K}},
\textbf{V}_s=\textbf{X}_{(l \times d)}\textbf{W}_{\mathrm{V}},
\end{equation}
\begin{equation}
\label{eqn: spatial attention 1}
\textbf{Z}_s = \mathrm{Softmax}(\textbf{Q}_s\textbf{K}_s^T/\sqrt{d_{\mathrm{k}}}+\textbf{B})\textbf{V}_s
\end{equation}

Spatial Attention Module can be formulated as:
\begin{align}
\label{eqn: spatial attention}
& \textbf{Z}_s=\mathrm{SA}(\mathrm{LN}(\textbf{X})) + \textbf{X},\\
& \textbf{O}_s=\mathrm{MLP}(\mathrm{LN}(\textbf{Z}_s)) + \textbf{Z}_s,
\end{align}
where \(\textbf{X}_{(l \times d)}\) denotes the embedding of patches, \(l\) represents the number of tokens, and \(d\) (\(d_{\mathrm{k}}\)) is the token dimension. \(\textbf{Q}\), \(\textbf{K}\), \(\textbf{V}\), which are the query matrix, key matrix, and value matrix. \(\textbf{b}\) is the relative position attention bias \cite{liu2021swin}. \(\mathrm{LN}\) is the layer normalization, and \(\mathrm{MLP}\) is a two-layer non-linear fully connected neural network.

To enhance the sensitivity of spatial attention to photometric abnormalities, we propose Photometric Consistency Loss (PCL), designed to increase the similarity among frames of the same video. The loss function is defined as: 
\begin{equation}
\tiny
\label{eqn: PCL}
\mathcal{L}_{\mathrm{PCL}}=
\begin{cases}
    -\frac{1}{BT}\sum_i^{BT} \sum_{\substack{j \\j \neq i \\ y_i=y_j}}^{BT} 
\frac{\exp(s_{i,j}/\tau)}{\sum_{\substack{k \\ k \neq i}}^{BT} \exp(s_{i,k}/\tau)},  & y = \mathrm{1}  \\

  -\frac{1}{BT}\sum_i^{BT} \sum_{\substack{j \\j \neq i \\ y_i=y_j}}^{BT} \mathbf{I}(\lfloor \frac{i}{T} \rfloor= \lfloor \frac{j}{T} \rfloor)
  \frac{\exp(s_{i,j}/\tau)}{\sum_{\substack{k \\ k \neq i}}^{BT} \exp(s_{i,k}/\tau)},  & y = \mathrm{0}
\end{cases}
\end{equation}
where  $\tau$ is a temperature parameter, $y$ represents the liveness label (1 is live, 0 is spoof). $s_{i,j}$ is the cosine similarity between $i$-th class token and $j$-th class token. $B$ denotes the batch size, $T$ denotes the number of frames. $\lfloor \frac{i}{T} \rfloor$ represents round down.

Notably, live and spoof samples are treated asymmetrically in PCL. For live samples, we bring all live frames from different videos closer in the representation space, reflecting their inherent photometric homogeneity. In contrast, for spoof samples, only frames within the same video are clustered together, as spoof samples tend to exhibit significant photometric heterogeneity across videos.

\subsection{Topology-aware Spatial Attention}
\label{subsec:spatial attention}
Facial landmarks extracted from videos are naturally represented as spatiotemporal graph data. However, transformers inherently have a global receptive field, enabling unrestricted information aggregation between any two tokens. While advantageous in some contexts, this innate property may lead to ambiguity regarding topology information. especially for graph-structured data, which lacks a sequential structure or 2D inductive bias inherent to images. Inspired by Graphormer \cite{ying2021transformers}, we employ topology-aware spatial attention for spatiotemporal graph representation learning.

In an undirected graph, the adjacency matrix encodes the topological relationships among nodes. Fortunately, the intrinsic topology capacity of the adjacency matrix can be achieved by an attention mask that can reflect the topology. Specifically, if there is no connection between two nodes, their attention weight (\(\textbf{Q}\textbf{K}^T(i,j)\)) is set to negative infinity, signifying a lack of direct information interaction. Or when the number of minimal hops between two nodes is too long, the exchange of information is also limited. To this end, we define a function \(\phi_h(n_i,n_j)\) to express the adjacency relationship between two nodes, if node \(n_i\) cannot reach node \(n_j\) within a maximum of \(h\) hops, it is considered that there is no connection between \(n_i\) and \(n_j\). Otherwise, the shortest number of hops between \(n_i\) and \(n_j\) serves as the relative distance between them:
\begin{equation}
\label{eqn: spatial pos encoding}
\small
\phi_h(n_i,n_j) = 
\begin{cases}
\mathrm{Embed}(\mathrm{hop_{min}}(n_i,n_j)),&\mathrm{hop_{min}}(n_i,n_j) \leq h \\
-\mathrm{inf},& \mathrm{otherwise}
\end{cases}
\end{equation} 
\begin{equation}
\label{eqn: positional embedding}
B_{\mathrm{graph}}(i,j) = \phi_h(n_i,n_j),
\end{equation}
where the output of function \(\phi_h(n_i,n_j)\) is a learnable real value, and the \(\mathrm{hop_{min}}(n_i,n_j)\) represents the minimum hop from \(n_i\) to \(n_j\), returning an index that corresponds to the learnable real value. The sketch of landmarks topology is shown in Fig. \ref{fig: Spatial Temporal Forms} (b).

In addition to encoding topological relationships, nodes in a graph vary in their importance, a property known as node centrality or degree. To incorporate this information, we add the centrality embedding, which is associated with degree, to the feature of node embedding:
\begin{equation}
\label{eqn: centrality encoding}
h_{n_i} = \mathrm{NodeEmbed}(\gamma(x_{n_i})) + \mathrm{CentralEmbed}(n_i),
\end{equation}
where \(x_{n_i}\) is the original coordinates of landmarks, \(h_{n_i}\) denotes the feature of node with centrality embedding. The centrality embedding serves as a positional embedding for graph data, where the 'position' information reflects whether a node is located in the central region of the graph. Specifically, we apply a lookup table to store the learnable centrality embeddings corresponding to various node degrees, ensuring that nodes with the same degree share the same centrality embedding. By combining topology-aware spatial attention with node centrality embedding, model can effectively captures the structure of graph-structured data.

\subsection{Kronecker Temporal Attention}
\label{subsec:temporal attention}
The pipeline temporal attention cannot ensure interested objects always appear in patches at the same spatial location, thereby limiting the scope of capturing dynamic clues, due this form of temporal receptive field is constrain within a time pipeline, we refer to it as pipeline temporal attention. This may lead to the model deviating in their understanding of information, especially in tasks such as face anti-spoofing. To address these drawbacks, we introduce the Kronecker temporal attention. Specifically, a patch at timestamp \(t\) can interact with all other patches, excluding those sharing the same timestamp \(t\). It will increase the search range (wider temporal receptive field) of each patch in temporal dimension and can adaptively capture favorable information. Compared to the combination of pipeline temporal attention and spatial attention, the combination of Kronecker temporal attention and spatial attention has better spatiotemporal complementarity. This attention mode can be achieved through the joint attention employed a mask. The trick for obtaining the temporal attention mask is using the Kronecker product. Hence, we term it Kronecker temporal attention as illustrated in Fig. \ref{fig: Spatial Temporal Forms} (c).

Kronecker Temporal Attention (KTA) can be formulated as:
\begin{equation}
\label{eqn: temporal attention 0}
\textbf{Q}_t=\textbf{X}_{(tl \times d)}\textbf{W}_{\mathrm{Q}},
\textbf{K}_t=\textbf{X}_{(tl \times d)}\textbf{W}_{\mathrm{K}},
\textbf{V}_t=\textbf{X}_{(tl \times d)}\textbf{W}_{\mathrm{V}} ,
\end{equation}
\begin{equation}
\label{eqn: Kronecker mask}
\textbf{M}_{(tl \times tl)} = [\textbf{I}_{(t \times t)} \otimes (\textbf{J}_{(m \times m)}-\textbf{J}_{(m \times m)})]_{1 == -\mathrm{inf}},
\end{equation}
\begin{equation}
\label{eqn: temporal attention 1}
\textbf{Z}_t = \mathrm{Softmax}(\textbf{Q}_t\textbf{K}_t^T/\sqrt{d_{\mathrm{k}}}+\textbf{M})\textbf{V}_t
\end{equation}

Kronecker Temporal Attention Module can be formulated as:
\begin{align}
\label{eqn: temporal attention}
\
& \textbf{Z}_t = \mathrm{KTA}(\mathrm{LN}(\textbf{X})) + \textbf{X},\\
& \textbf{O}_t = \mathrm{MLP}(\mathrm{LN}(\textbf{Z}_t)) + \textbf{Z}_t,
\end{align}
where \(\textbf{X}_{(tl \times d)}\) denotes the embedding of patches or nodes from \(t\) frames, \(tl\) represents the number of tokens from \(t\) frames and \(d\) (\(d_{\mathrm{k}}\)) is the token dimension. \(\textbf{I}\) represents an identity matrix, \(\textbf{J}\) is the all-ones matrix, \([\quad]_{1==-\mathrm{inf}}\) means replacing ones in the matrix with negative infinity, \(\textbf{M}\) is the temporal attention mask.

For graph temporal attention, we also apply the same Kronecker temporal mask to enhance the ability to capture landmark motion. For each node, in the graph temporal attention, except for other nodes in the current frame that are not visible, all nodes in other frames are visible to it, which matches the vision stream temporal attention. The purpose of doing this is twofold: firstly, to introduce the advantage of expanding the temporal receptive field into graph attention; secondly, to lay the foundation for using graph temporal attention to guide vision stream temporal attention.

\section{Graph Guided Vision Temporal Attention}
\label{subsec:fusion of statics and dynamics}
\begin{figure}[t!]
    \centering
    \includegraphics[width=0.49\textwidth,height=0.17\textwidth]{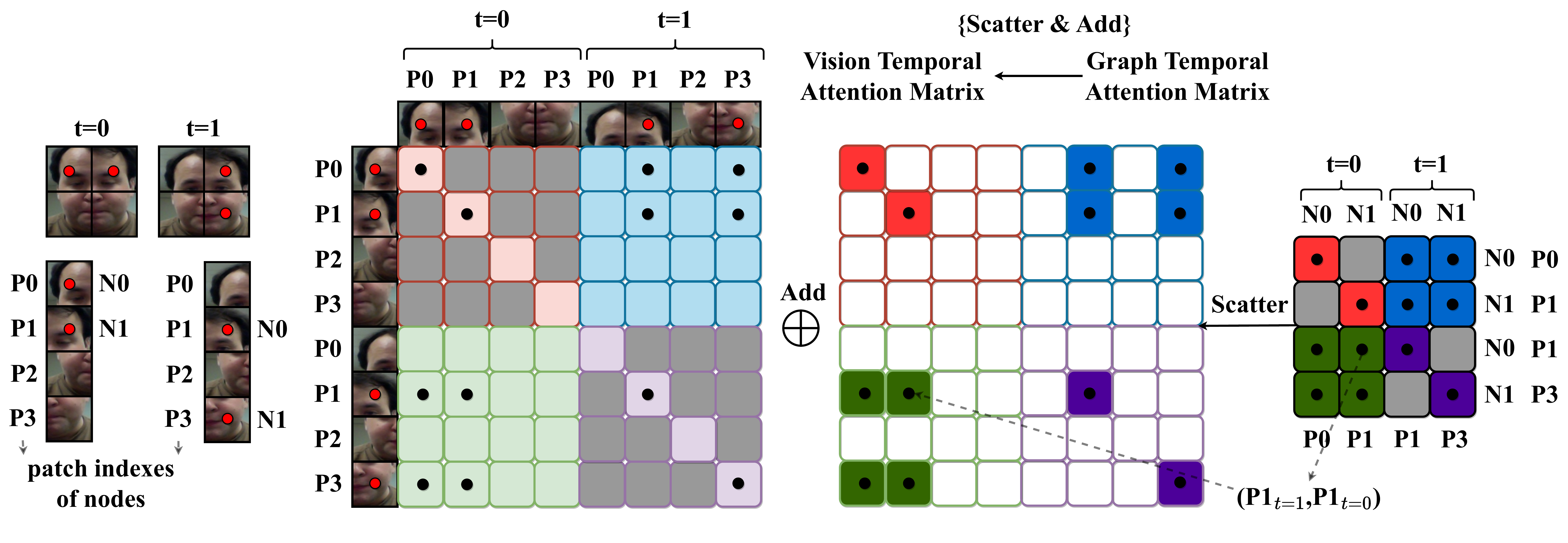}
    \vspace{-0.5cm}
    \caption{Graph-guided vision temporal attention. First, performing a scatter operation to align the shape of the graphic temporal attention matrix with that of the visual temporal attention matrix, and then adding them together.}
    \label{fig: Scatter Add Attention}
\vspace{-0.3cm}
\end{figure}

The video datasets from face anti-spoofing exhibit significant differences from general video datasets attributed to the similarity in facial structures (high content homogeneity) and the subtle, inconspicuous motion of faces in videos. Although the Kronecker temporal attention can compensate for the shortcomings of limited temporal receptive field, focusing on a wider scope to search for dynamic clues, but it may struggle to quickly locate key motion areas. Hence, we hope the temporal attention promptly focuses on more useful details. Facial landmarks can help identify the facial micro-motions that reveal differences between live and spoof motions, such as the motion in the corner of the mouth and eye blinking, which are all potential regions of interest. In Fig. \ref{fig: Scatter Add Attention}, each landmark belongs to a patch box, and the motion trajectory of landmarks can express facial motion to an extent. Thus, we naively believe patch boxes that contain landmarks are most likely to undergo pixel changes, and the motion of landmarks helps guide the capture of pixel motion.

During fine-tuning, we utilize the sequence number of the patches as indexes where the landmark node is located. Nodes are scattered based on their patch sequence number indexes. After performing a scatter operation, the shape of the graph temporal attention matrix is aligned with the shape of the vision temporal attention matrix, and then adding them. In Fig. \ref{fig: Scatter Add Attention}, we provide an example demonstration with a patch number of 4 and a node number of 2. At $t$=0, N$_0$ belongs to the 0-th patch P$_0$, N$_1$ belongs to the 1-th patch P$_1$, while at $t$=1, N$_0$ belongs to the P$_1$, and N$_1$ belongs to the P$_3$. In simpler terms, this involves replacing the row and column numbers of the graph temporal attention matrix with patch sequence numbers based on indexes. Subsequently, we select (scatter) and fill (add) elements. We term this operation scatter and add.

At each layer of the model, after executing the spatial attention of the graph and vision, we first calculate the graph Kronecker temporal attention and cache its attention matrix. During the subsequent vision Kronecker temporal attention, the cached attention matrix is scattered and added as an attention bias, guiding the dynamic information capture for the vision stream:
\begin{equation}
\textbf{B}_{\mathrm{guide}} = \textbf{Q}_{gt}\textbf{K}_{gt}^T/\sqrt{d_\mathrm{k}},
\end{equation}
\begin{equation}
\label{eqn: vision temporal attention}
\small
\textbf{Z}_{vt} = \mathrm{Softmax}(\textbf{Q}_{vt}\textbf{K}_{vt}^T/\sqrt{d_\mathrm{k}}+\textbf{M}+\mathrm{ScatterAdd}(\textbf{B}_{\mathrm{guide}}))\textbf{V}_{vt},
\end{equation}
where \(B_{\mathrm{guide}}\) is graph guidance bias. The subscript \(gt\) represents the graph temporal attention, and \(vt\) represents the vision temporal attention.

\section{Experiment}
\label{sec:experiment}
\subsection{Datasets and Protocol}
\label{subsec:datasets and protocol}
We use nine public datasets: Oulu-NPU \cite{boulkenafet2017oulu} (denoted as O), CASIA-FASD \cite{zhang2012face} (denoted as C), Idiap Replay-Attack \cite{chingovska2012effectiveness} (denoted as I), MSU-MFSD \cite{wen2015face} (denoted as M), CASIA-SURF \cite{zhang2019dataset,zhang2020casia}, CASIA-CeFA \cite{liu2021casia,liu2021cross}, WMCA \cite{george2019biometric}, 3DMAD \cite{erdogmus2013spoofing}, and HKBU-MARsV1\cite{liu20163d}. We evaluate our model on the prevalent cross-dataset protocol to verify its effectiveness. To demonstrate the robust generalization capabilities of G$^2$V$^2$former, we assess its performance under the protocol of cross-dataset (3DM, HKM), cross-type (unseen attacks, 3D Mask). The evaluation metrics are Half Total Error Rate (HTER) and Area Under the Curve (AUC).

\subsection{Implementation Details}
\label{subsec:implementation details}
First, we extract all frames from each video and utilize the MTCNN \cite{zhang2016joint} for face detection, then crop and resize the images to 224 \(\times\) 224 for facial landmarks detection. We employ the 3DDFA-V2 \cite{guo2020towards} to obtain the 3-dimensional coordinates of landmarks. We apply random blurring, sharping, 2D rotation, and affine transformation for data augmentation. It is worth noting that 2D rotation and affine transformation of image are also applied to landmarks. We set the patch size to 16, and the length of the input frames is 8. The sample interval depends on the total frames (default 8). When the total length is shorter, we try to reduce the sampling interval until the requirements are met. We use Adam optimizer \cite{kingma2014adam}, and the base learning rate is set to 5e-4 in pre-training, 1e-4 in fine-tuning, The scheduler for reducing learning rates is cosine decay\cite{loshchilov2016sgdr}, and the total pre-training iterations are 60000, 20000 iterations for fine-tuning. The encoder and decoder adopt an asymmetric design, with 12 layers for the encoder and 4 layers for the decoder. Our method is implemented under the Pytorch framework.

\noindent \textbf{The detailed Architecture of G$^2$V$^2$former.}
We use an asymmetric encoder-decoder architecture for video self-supervised pre-training and discard the decoder during fine-tuning. We take the TimeSformer \cite{bertasius2021space} equipped with Kronecker temporal attention as our backbone. The specific architecture is depicted in Tab. \ref{table: architecture}.

\noindent \textbf{The Detailed Training Settings.}
We load the weight of ViT for spatial attention module initialization. We list the pre-training and fine-tuning configuration in Tab. \ref{table: pre-training setting} and \ref{table: fine-tuning setting}, with detailed parameter settings for reference. The setting is the same in all protocols.

\noindent \textbf{Frame Sampling Strategy.}
Our frame sampling strategy driven by the observation that facial videos in anti-spoofing datasets often contain highly repetitive information. Therefore, it is unwise to continuously extract numerous frames of a video at a sampling interval of 1. We have attempted to use consecutive 16, 32 even 64 frames as input, but the computational overhead involved is formidable, due to the computational complexity of temporal attention calculation is \(O(t^2l^2)\), where \(t\) is the length of frames, \(l\) is the token amount in each frame. Training with larger frame lengths, such as 64 frames, became time-consuming and eventually unsustainable. So we need to cover as long a time span as possible and bring as little computational overhead as possible.

To this end, we design a sampling rule that randomly selects a starting point from a complete video, sample 8 frames with a sampling interval of 4 (or 8) frames. When the total length of the video is less than 4 (8) \(\times\) 8, we try to reduce the sampling interval to 3 (6), 2 (4) until the requirements are met. This ensures that we cover a long time span with minimal computational overhead. By adopting this sampling strategy, we only need to sample 8 frames from a video to have a high probability of encountering dynamic facial expressions in long-distance videos. This sampling strategy effectively solves the encountered challenges. We use 8-frame segments for training. During testing, we extract one or more 8-frame segments from a long video and average the results.

\begin{table}[t!]
\caption{Architecture, where T denotes the number of frames, N denotes the number of tokens, and C denotes the number of channels. L is the number of frequencies from positional encoding.}
\label{table: architecture}
\centering
\setlength{\arrayrulewidth}{0.8pt}
\resizebox{0.48\textwidth}{!}{
\begin{tabular}{c|c|c}
\hline
\textbf{Module} & \textbf{Block} & \textbf{Output Size} (T$\times$N$\times$C)\\
\hline
\multirow{2}{*}{input data} & frames & 8 $\times$ 3 $\times$ 224 $\times$ 224  \\
                                                          & landmarks &  8 $\times$ 68 $\times$ 3 $\times$ 2$^L$ \\ 
\hline
\multirow{2}{*}{encoder} &  vision $ \left[
\begin{array}{l} 
\text{SA(768)}  \\
\text{KTA(768) } 
\end{array}
\right] $$\times$ 12& 8 $\times$ 196 (+1) $\times$ 768\\ 
                                                    & graph $ \left[
\begin{array}{l} 
\text{SA(128)}  \\
\text{KTA(128) } 
\end{array}
\right] $$\times$ 12& 8 $\times$ 68 (+1) $\times$ 128\\ 
\hline
\multirow{2}{*}{decoder} &  vision $\left[
\begin{array}{l} 
\text{SA(384)}  \\
\text{KTA(384) } 
\end{array}
\right]$ $\times$ 4& 8 $\times$ 196 $\times$ 384\\ 
                                                     & graph $\left[
\begin{array}{l} 
\text{SA(64)}  \\
\text{KTA(64) } 
\end{array}
\right]$ $\times$ 4& 8 $\times$ 68 $\times$ 64\\ 
\hline
\multirow{2}{*}{projector} & vision $\left[\begin{array}{l} 
\text{MLP(768) } \end{array}\right]$
& 8 $\times$ 3 $\times$ 224 $\times$ 224 \\ 
                                                        & graph $\left[\begin{array}{l} 
\text{MLP(3 $\times$ 2$^L$) } \end{array}\right]$  
& 8 $\times$ 68 $\times$ 3 $\times$ 2$^L$ \\ 
\hline
classifier & vision cat graph $\left[\begin{array}{l} 
\text{MLP(2) } \end{array}\right]$ & 8 $\times$ 2 \\ 
\hline
\end{tabular}}
\end{table}

\begin{table}[t!]
\vspace{-0.3cm}
\caption{Pre-training setting}
\label{table: pre-training setting}
\centering
\setlength{\arrayrulewidth}{0.8pt}
\resizebox{0.5\textwidth}{!}{
\begin{tabular}{p{5cm}p{5cm}}
\hline
\textbf{Config}& \textbf{Param}\\
\hline
optimizer & AdamW \\ 
base learning rate & 5e-4\\
weight decay & 0.05 \\
optimizer momentum & $\beta_1$, $\beta_2$ = 0.9, 0.95\\
batch size & 4 \\
learning rate schedule & Cosine decay \\
warmup iterations& 4000\\
training iterations& 60000\\
augmentation & Rotation, Affine, Blur, Sharpness \\
seed & 666 \\
\hline
\end{tabular}
}
\end{table}

\begin{table}[t!]
\vspace{-0.3cm}
\caption{Fine-tuning setting}
\label{table: fine-tuning setting}
\centering
\setlength{\arrayrulewidth}{0.8pt}
\resizebox{0.5\textwidth}{!}{
\begin{tabular}{p{5cm}p{5cm}}
\hline
\textbf{Config}& \textbf{Param}\\
\hline
optimizer & AdamW \\
base learning rate & 1e-4\\
weight decay & 0.05 \\
optimizer momentum & $\beta_1$, $\beta_2$ = 0.9, 0.999\\
batch size & 2\\
learning rate schedule & Cosine decay \\
warmup iterations& 500\\
training iterations& 20000\\
augmentation & Blur, Sharpness \\
label smoothing & 0.1 \\
dropout & 0.05\\
drop path & 0.05\\
layer-wise lr decay & 0.75 \\
\hline
\end{tabular}
}
\vspace{-0.3cm}
\end{table}

\begin{table*}[t!]
\caption{Comparison to state-of-the-art FAS method under the LOO setting. The bold numbers indicate the SOTA, star marks (*) the video-based methods. Others are single-frame-level methods.}
\label{table: OCMI}
\centering
\setlength{\arrayrulewidth}{0.8pt}
\resizebox{0.9\textwidth}{!}{
\begin{tabular}{ccccccccc}
\hline
\multirow{2}{*}{\textbf{Method}} & \multicolumn{2}{c}{\textbf{O\&C\&I to M}} & \multicolumn{2}{c}{\textbf{O\&M\&I to C}} & \multicolumn{2}{c}{\textbf{O\&C\&M to I}} & \multicolumn{2}{c}{\textbf{I\&C\&M to O}} \\
\cmidrule(r){2-3} \cmidrule(r){4-5} \cmidrule(r){6-7} \cmidrule(r){8-9}
& HTER(\%) & AUC(\%) & HTER(\%) & AUC(\%) & HTER(\%) & AUC(\%) & HTER(\%) & AUC(\%) \\
\hline
MMD-AAE \cite{li2018domain} & 27.08 & 83.19 & 44.59 & 58.29  & 31.58 & 75.18 & 40.98 & 63.08 \\
MADDG \cite{shao2019multi} & 17.69 & 88.06 & 24.50 & 84.51 & 22.19 & 84.99 & 27.98 & 80.02 \\
NAS-FAS \cite{yu2020fas} & 19.53 & 88.63   & 16.54 & 90.18  & 14.51 & 93.84  & 13.80 & 93.43 \\
RFM \cite{shao2020regularized} & 13.89 & 93.98   & 20.27 & 88.16  & 17.30 & 90.48  & 16.45 & 91.16 \\
SSDG \cite{jia2020single} & 7.38  & 97.17  & 10.44  & 95.94 & 11.71  & 96.59 & 15.61  & 91.54 \\
D\(^2\)AM \cite{chen2021generalizable} & 12.70  & 95.66  & 20.98  & 85.58 & 15.43  & 91.22 & 15.27  & 90.87 \\
DRDG \cite{liu2021dual} & 12.43  & 95.81  & 19.05  & 88.79 & 15.56  & 91.79 & 15.63  & 91.75 \\
ANRL \cite{liu2021adaptive} & 10.83  & 96.75  & 17.83  & 89.26 & 16.03  & 91.04 & 15.67  & 91.90 \\
SSAN \cite{wang2022domain} & 6.67  & 98.75  & 10.00  & 96.67 & 8.88  & 96.79 & 13.72  & 93.63 \\
PatchNet \cite{wang2022patchnet} & 7.10  & 98.46  & 11.33  & 94.58 & 13.40  & 95.67 & 11.82  & 95.07 \\
IADG \cite{zhou2023instance} & 5.41  & 98.19  & 8.70  & 96.44 & 10.62  & 94.50 & 8.86  & 97.14 \\
SA-FAS \cite{sun2023rethinking} & 5.95  & 96.55  & 8.78  & 95.37 & 6.58 & 97.54 & 10.00  & 96.23 \\
UDG-FAS \cite{liu2023towards} & 5.95 & 98.47 & 9.82 & 96.76 & 5.86 & 98.62 & 10.97  & 95.36 \\
TTDG \cite{zhou2024test} & 7.91 & 96.83 & 8.14 & 96.49 & 6.50 & 97.98 & 10.00  & 95.70 \\
GAC-FAS \cite{le2024gradient} & 5.00 & 97.56 & 8.20 & 95.16 & \textbf{4.29} & 98.87 & 8.60  & \textbf{97.16} \\
\hline
Auxiliary \cite{liu2018learning}* & 22.72  & 85.88 & 33.52 & 73.15 & 29.14 & 71.69 & 30.17  & 77.61 \\ 
CCDD\cite{saha2020domain}* & 15.42  & 91.13 & 17.41 & 90.12 & 15.87 & 91.72 & 14.72  & 93.08 \\ 
GAIN \cite{chang2023closer}* & 4.05  & 98.92  & 8.52 & 96.02 & 8.50  & 97.27 & 12.50  & 95.12 \\ 
\hline
\textbf{G$^2$V$^2$former}* & \textbf{3.72}  & \textbf{99.24}  & \textbf{5.59}  & \textbf{97.64} & 5.42  & \textbf{98.95} & \textbf{8.21} & 97.04 \\
\hline
\end{tabular}
}
\vspace{-0.2cm}
\end{table*}

\begin{table*}[t!]
\caption{Evaluation on cross-domain protocols among CASIA-SURF (S), CASIA-CeFA (C), WMCA (W)}
\label{table: WCS}
\centering
\setlength{\arrayrulewidth}{0.8pt}
\resizebox{0.85\textwidth}{!}{
\begin{tabular}{cccccccc}
\hline
& \multirow{2}{*}{\textbf{Method}} & \multicolumn{2}{c}{\textbf{CS $\rightarrow$ W}} & \multicolumn{2}{c}{\textbf{SW $\rightarrow$ C}} & \multicolumn{2}{c}{\textbf{CW $\rightarrow$ S}} \\
\cmidrule(r){3-4} \cmidrule(r){5-6} \cmidrule(r){7-8}
& & HTER $\downarrow$ (\%)& AUC $\uparrow$ (\%)& HTER $\downarrow$ (\%)& AUC $\uparrow$ (\%)& HTER $\downarrow$ (\%)& AUC $\uparrow$ (\%)\\
\hline
\multirow{3}{*}{0-shot} & SSDG \cite{jia2020single}& 12.64 & 94.35 & 12.25 & 94.78 & 27.08 & 80.05 \\
& ViT \cite{huang2022adaptive} & 7.98 & 97.97 & 11.13 & 95.46 & 13.35 & 94.13 \\
& \textbf{G$^2$V$^2$former} & \textbf{6.21}& \textbf{98.56}& \textbf{10.03}& \underline{95.42}& \textbf{12.66}& \textbf{94.17}\\
\hline
\multirow{4}{*}{5-shot} & SSDG \cite{jia2020single}& 5.08 & 99.02 & 6.72 & 98.11 & 18.88 & 88.25 \\
& ViT \cite{huang2022adaptive} & 4.30 & 99.16 & 7.69 & 97.66 & 12.26 & 94.40 \\
& ViTAF \cite{huang2022adaptive} & 4.51 & 99.44 & 7.21 & 97.69 & 11.74 & 94.13 \\
& ViTAF* \cite{huang2022adaptive} & 2.91 & 99.71 & 6.00 & 98.55 & 11.60 & 95.03 \\
& \textbf{G$^2$V$^2$former} & \textbf{2.59} & 99.20 & \textbf{4.99} & \textbf{98.62} & \textbf{9.58} & \textbf{95.60}\\
\hline
\end{tabular}
}
\vspace{-0.3cm}
\end{table*}

\subsection{Comparison with SOTA Methods}
\label{subsec:comparison with sota methods}
We evaluate our model under the Leave-One-Out (LOO) setting, which utilizes O, C, I, and M datasets. Besides, in order to demonstrate our model's generalization of unseen attacks, we also used two additional datasets of 3D mask attacks for testing.

\noindent \textbf{Leave-One-Out (LOO).}
In the LOO setting, we randomly select three datasets out of O, C, I, and M for training, while the remaining one is reserved for testing. These four datasets contain the same type of face samples (live, printed photo, video replay), but they encounter variations in capture device, illumination, resolution, etc. As depicted in Tab. \ref{table: OCMI}, our method exhibits superior performance when compared with both frame-level methods and video-level methods (marked with a star). This reveals the significance of spatiotemporal fusion in face anti-spoofing tasks.

\noindent \textbf{Evaluation on CASIA-SURF, CASIA-CeFA and WMCA Protocol.} As shown in Tab. \ref{table: WCS}, we evaluate our method among CASIA-SURF, CASIA-CeFA and WMCA. They all exhibit rich and diverse variations of expression and motion. We follow the experimental setting in \cite{huang2022adaptive}. Under the 0-shot and 5-shot scenarios, our G$^2$V$^2$former surpasses most previous methods. Due to our utilization of both photometric and dynamic information, allow us to more fully mine spoofing cues.

\noindent \textbf{Limited Source Domain Protocol}. As shown in Tab. \ref{table: MI}, we evaluate our method under the limited source domains. To ensure a fair comparison, we follow the previous work's settings, which select M and I as source domains, while C and O, are respectively utilized as the target domain. In scenarios with limited source data, G$^2$V$^2$former still largely surpasses most previous methods. Spatiotemporal fusion exhibits superior performance when compared with frame-level methods. This reveals the significance of spatiotemporal fusion in face anti-spoofing tasks.

\noindent \textbf{Generalization of Unseen Attacks.} 
The OCIM-LOO primarily consists of printed photos and video replay, focusing on the cross-dataset intra-type protocol which necessitates overcoming domain gaps. In contrast, the cross-dataset and cross-type protocols are more challenging. Defending 3D mask attacks, given their high degree of photorealism, demands specialized defense methods. However, defending against 3D mask attacks becomes challenging in the absence of relevant training samples. To explore the strengths of G$^2$V$^2$former in more demanding scenarios, we propose two cross-dataset cross-type protocols: O\&C\&I to 3DM, and O\&C\&I to HKM. The 3DM (3DMAD \cite{erdogmus2013spoofing}), and HKM (HKBU-MARsV1 \cite{liu20163d}) involve low, high-photorealism custom wearable masks. For comparison, we select a representative photometric-based algorithm SSDG \cite{jia2020single} and vanilla ResNet18 \cite{he2016deep}. As depicted in Table \ref{table: Generalization}, SSDG experienced a significant performance drop, ResNet18 is vulnerable or even ineffective, while our method still holds on. This indicates that the G\(^2\)V\(^2\)former has an unexpected generalization ability. The performance on HKBU-MARsV1 is slightly inferior because its 3D mask has higher photorealism.

\begin{table}[t!]
\caption{Experiments on limited source domains setting.}
\label{table: MI}
\centering
\setlength{\arrayrulewidth}{0.8pt}
\resizebox{0.49\textwidth}{!}{
\begin{tabular}{cccccc}
\hline
\multirow{2}{*}{\textbf{Method}} & \multicolumn{2}{c}{\textbf{M\&I to C}} & \multicolumn{2}{c}{\textbf{M\&I to O}} \\
\cmidrule(r){2-3} \cmidrule(r){4-5}
& HTER(\%) & AUC(\%) & HTER(\%) & AUC(\%)  \\
\hline
MADDG \cite{shao2019multi} & 41.02 & 64.33 & 39.35 & 65.10  \\
SSDG-M \cite{jia2020single} & 31.89  & 71.29  & 36.01  & 66.88 \\
D\(^2\)AM \cite{chen2021generalizable} & 32.65  & 72.04  & 27.70 & 75.36 \\
DRDG \cite{liu2021dual} & 31.28  & 71.50  & 33.35  & 69.14 \\
ANRL \cite{liu2021adaptive} & 31.06  & 72.12  & 30.73  & 74.10 \\
SSAN-M \cite{wang2022domain} & 30.00  & 76.20  & 29.44  & 76.62 \\
IADG \cite{zhou2023instance} & 23.51  & 84.20  & 22.70  & 84.28 \\
TTDG \cite{zhou2024test} & 19.67 & 85.44  & 18.70  & 90.09 \\
GAC-FAS \cite{le2024gradient} & 20.71  & 84.29  & 19.71  & 89.01 \\
\hline
\textbf{G$^2$V$^2$former} & \textbf{18.43}  & \textbf{87.64}  & \textbf{17.47}  & 88.97 \\
\hline
\end{tabular}
}
\end{table}

\begin{table}[t!]
\vspace{-0.3cm}
\caption{Generalization evaluation on 3D mask attacks. Experiments are conducted on two popular 3D mask benchmarks, 3DMAD (denoted as 3DM) and HKBU-MARsV1 (denoted as HKM).}
\label{table: Generalization}
\centering
\setlength{\arrayrulewidth}{0.8pt}
\resizebox{0.49\textwidth}{!}{
\begin{tabular}{ccccc}
\hline
& \multicolumn{2}{c}{\textbf{O\&C\&I to 3DM}} & \multicolumn{2}{c}{\textbf{O\&C\&I to HKM}} \\
\cmidrule(r){2-3} \cmidrule(r){4-5}
& HTER(\%) & AUC(\%) & HTER(\%) & AUC(\%) \\
\hline
ResNet18 \cite{he2016deep}& 48.75 & 61.07  & 50.00  & 59.40 \\
SSDG \cite{jia2020single}& 37.07& 72.93& 46.00& 65.20\\
IADG \cite{zhou2023instance}& 22.04& 80.39& 28.11& 69.20\\
SA-FAS \cite{sun2023rethinking} & 23.75& 81.66& 30.47& 71.54\\
\hline
\textbf{G$^2$V$^2$former} & \textbf{1.26}& \textbf{99.69}& \textbf{9.34}& \textbf{95.52}\\
\hline
\end{tabular}
}
\vspace{-0.2cm}
\end{table}

\section{Ablation Studies}
\label{subsec:ablation studies}
\noindent In this subsection, we conduct ablation studies to explore the efforts of each component. We also compare various spatiotemporal modules for spatiotemporal fusion with our spatial \& Kronecker temporal attention module. The results indicates that our Kronecker temporal attention is more suitable for FAS tasks. It should be noted that both the ablation study and comparison experiments are conducted under O\&C\&I\& to M settings.

\noindent \textbf{Contribution of Different Components.} 
Tab. \ref{table: Ablation} shows the ablation studies. For the baseline configuration, we employ the TimeSformer \cite{bertasius2021space} as the backbone equipped with spatial \& Kronecker temporal attention and extract the class token for classification. Then we introduce the graph stream whose input is facial landmarks. In this case, we naively concatenate the class token and hyper-node token for classification. Besides, we attempted to use the lower semantic coordinate motion of landmarks to guide vision temporal attention to capture higher semantic pixel motion over time. Finally, Photometric consistency loss is added to highlight the photometric characteristic in spatial attention. We can observe that under our baseline configuration, the performance has reached a relatively high value. After introducing landmarks without guidance, the performance has an improvement. When further exploring the exploitable value of the landmark, i.e., mining the high-level pixel motion semantic, points further increase. When introduce photometric consistency loss, our model reaches a peak in performance.

\begin{table}[t!]
\caption{Ablation of each component on O\&C\&I to M.}
\label{table: Ablation}
\centering
\setlength{\arrayrulewidth}{0.8pt}
\resizebox{0.48\textwidth}{!}{
\begin{tabular}{cccccc}
\hline
Baseline & Landmarks & Graph guide& PCL& HTER(\%) & AUC(\%) \\
\hline
\checkmark & - & - & - & 5.03  & 97.69 \\
\checkmark & \checkmark & - & - & 4.91 & 97.94 \\
\checkmark & \checkmark & \checkmark & - & \textbf{3.88} & \textbf{99.01} \\
\checkmark & \checkmark & \checkmark & \checkmark & \textbf{3.72}& \textbf{99.24}\\
\hline

\end{tabular}
}
\vspace{-0.3cm}
\end{table}

 \begin{table}[t!]
\caption{Comparison of different visual attention forms.}
\label{table: Comparison}
\centering
\setlength{\arrayrulewidth}{0.8pt}
\resizebox{0.49\textwidth}{!}{
\begin{tabular}{ccc}
\hline
Attention Form & HTER (\%) & AUC(\%) \\
\hline
3D Convolution (wo-guide)& 7.87& 95.65\\ 
1D+2D Convolution (wo-guide)& 7.90& 96.21\\ 
Joint Spatial-Temporal (wo-guide)& 6.33 & 96.86\\ 
Spatial-Temporal Pipeline (wo-guide)& 6.06 & 96.74\\
\textbf{Spatial-Kronecker Temporal (wo-guide)}& \textbf{4.91} & \textbf{97.94}\\
\hline
\end{tabular}
}
\vspace{-0.3cm}
\end{table}

\begin{table}[t!]
\centering
\caption{Training on variable frame length scenario, includes 1, 2, 4, 8. Inference at different fixed and variable frame length.}
\label{table: various frame length}
\setlength{\arrayrulewidth}{0.8pt}
\resizebox{0.46\textwidth}{!}{
\begin{tabular}{cccccc}
\hline
\multicolumn{6}{c}{\textbf{O\&C\&I to M} } \\
\hline
 \textbf{Frame length} & 1 & 2  & 4 & 8 & variable \\
\hline
\textbf{HTER} $\downarrow$ (\%) & 7.09 & 6.66 & 4.54 & \textbf{4.01} & 5.52\\
\textbf{AUC} $\uparrow$ (\%) & 97.31 & 97.49 & 98.46 & \textbf{99.12} & 98.80 \\
\hline
\end{tabular}
}
\end{table} 

\begin{figure}[t!]
\centering
    \includegraphics[width=0.45\textwidth,height=0.27\textwidth]{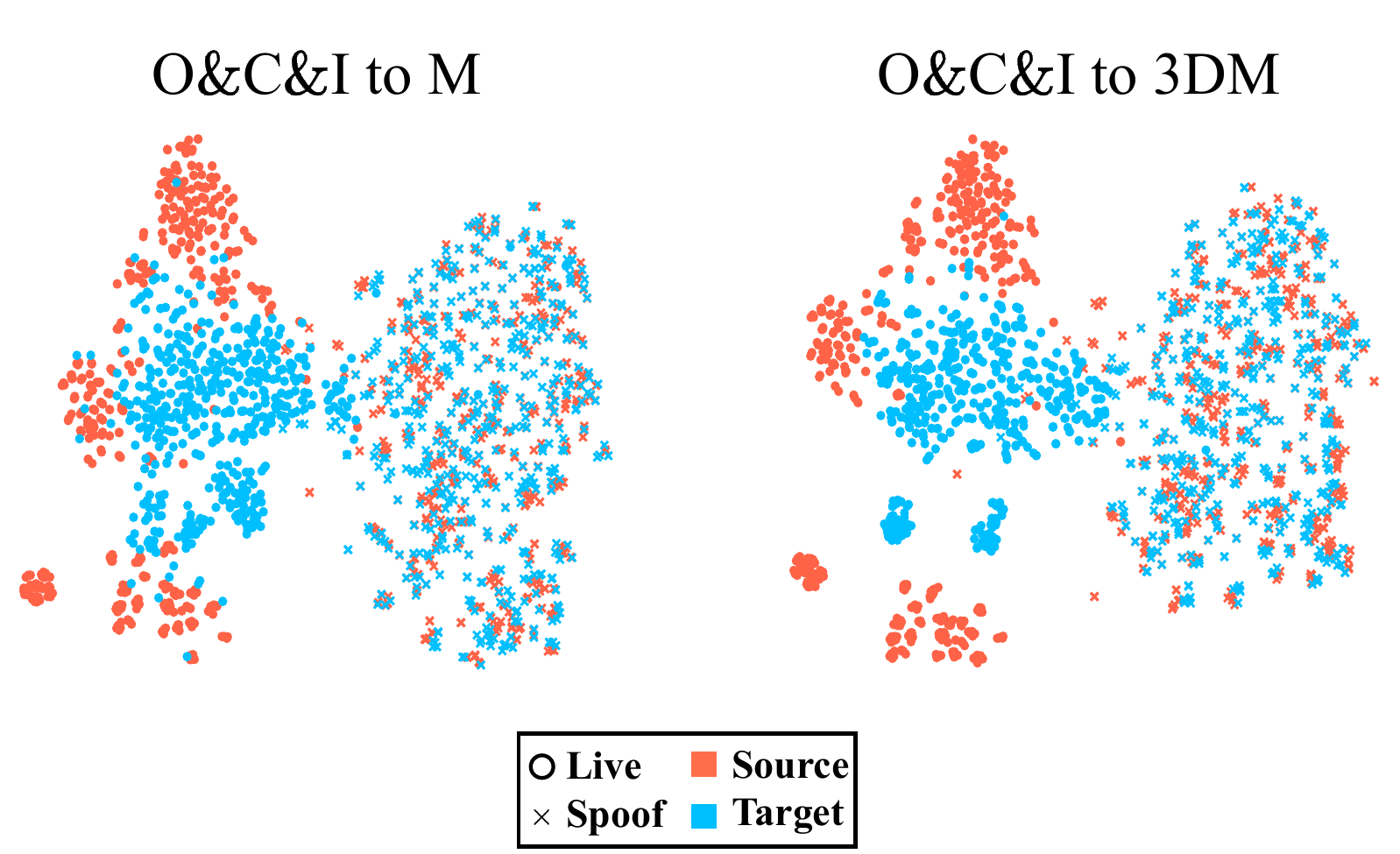}
    \caption{Feature distribution visualized by t-SNE. (Left) Cross-domain scenario, (Right) Cross-domain and unseen attack scenario.}
    \label{fig: TSNE}
\vspace{-0.3cm}
\end{figure}

\noindent \textbf{Comparison of Different Spatiotemporal Modules.} 
Tab. \ref{table: Comparison} depicts the comparison of different spatiotemporal modules. For the sake of fairness, we compared them without graphic guidance, due to the graphic guidance is incompatible with some of them. The results indicate that our method has advantages compared to the other spatiotemporal fusion module to the extent, and the Kronecker temporal attention is more beneficial for FAS tasks where pixel motion is not obvious.

\noindent \textbf{Variable Frame Length Scenarios.}
Our framework is compatible with variable frame length scenarios, because the shape of the Kronecker mask is determined by T and L. Therefore, we adopt a variable frame length sampling strategy during training, and employ a linear interpolation strategy for position encoding. During the inference, we test at 1, 2, 4, 8 frames, as well as variable frame length. As shown in the Table \ref{table: various frame length}, we conduct variable frame length training under the O\&C\&I to M protocol, then evaluate the performance at different fixed frame length settings. The results indicate that our framework also exhibit generalization ability in variable frame length scenarios.

\noindent \textbf{Inference time and Memory Usage.}
In Tab. \ref{table: Inference time and memory usage}, we show the inference time (s) and memory usage (GB) of G$^2$V$^2$former with different frame lengths and whether or not equip graph temporal guidance. \(8f\) denotes 8 frames, SA represents spatial attention, KTA represents Kronecker temporal attention,"wg", "wog" represent "with guide", "without guide", respectively.

\begin{table*}[t!]
\centering
\caption{Inference time \& memory usage on a single NVIDIA RTX 4090.}
\label{table: Inference time and memory usage}
\setlength{\arrayrulewidth}{0.8pt}
\resizebox{0.9\textwidth}{!}{
\begin{tabular}{cccccc}
\hline
\textbf{Torch Profiler}& SA-KTA\(_{8f}\)-wg& SA-KTA\(_{8f}\)-wog& SA-KTA\(_{4f}\)-wg& SA-KTA\(_{2f}\)-wg& SA-KTA\(_{1f}\)-wg\\
\hline
\textbf{Inference Time (s)} & 1.88 & 1.69 & 1.57 & 1.50 & 1.41\\
\textbf{Memory Usage (GB)} & 5.08 & 3.75 & 2.27 & 1.36 & 0.98\\
\hline
\end{tabular}
}
\end{table*}

\begin{figure*}[t!]
    \centering
    \includegraphics[width=0.99\textwidth,height=0.47\textwidth]{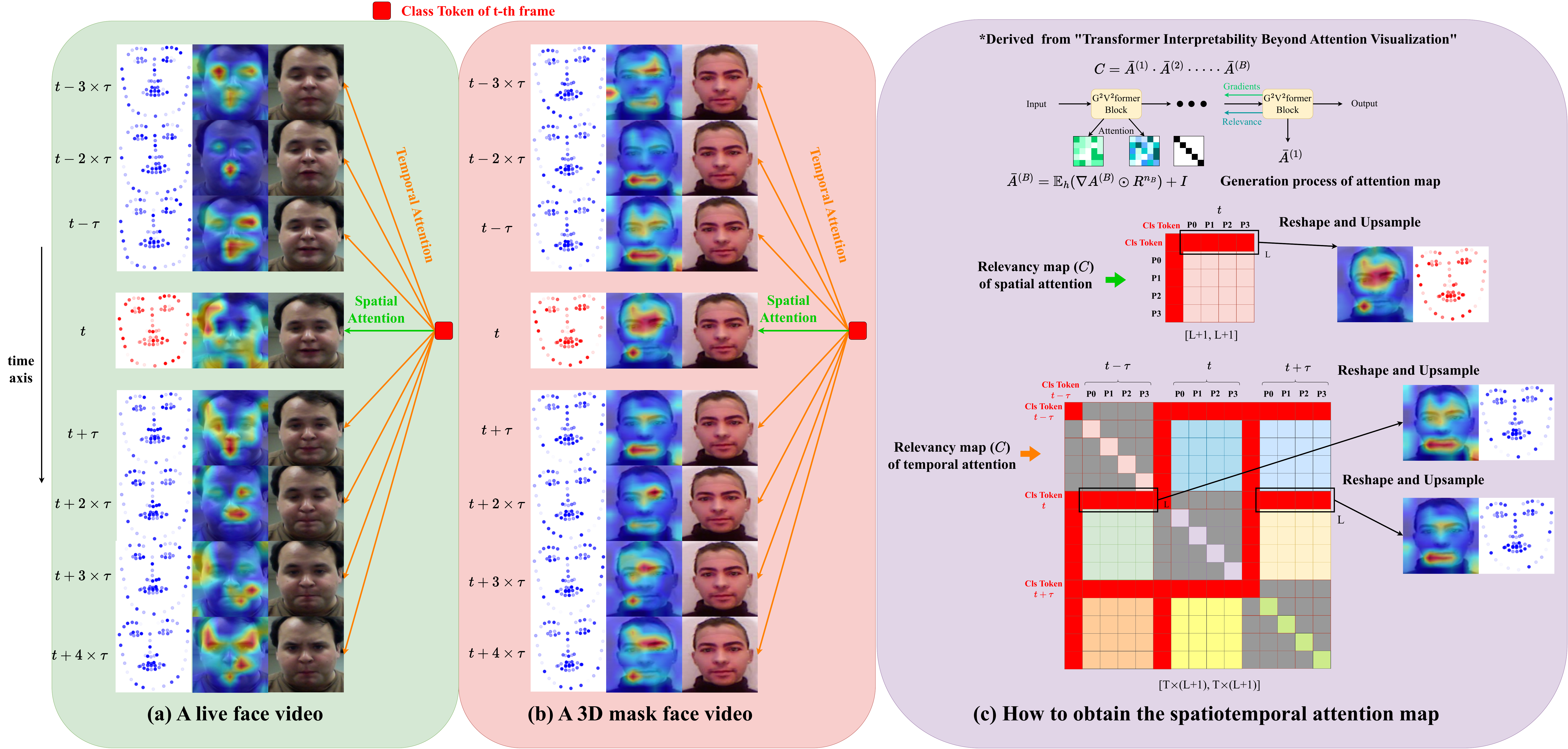}
    \caption{(a) and (b) Utilizing Transformer Explainability \cite{chefer2021transformer} to visualize a comprehensive spatiotemporal attention map. For vision branch, the transition from warm to cool colors corresponding to a shift from interested to uninterested. While in graph branch, the transition from opacity to transparency mirrors a change in attention values from high to low. (c) How to obtain the spatiotemporal attention map.}
    \label{fig: Explainability}
\end{figure*}

\begin{figure*}[t!]
    \centering
    \includegraphics[width=0.99\textwidth,height=0.57\textwidth]{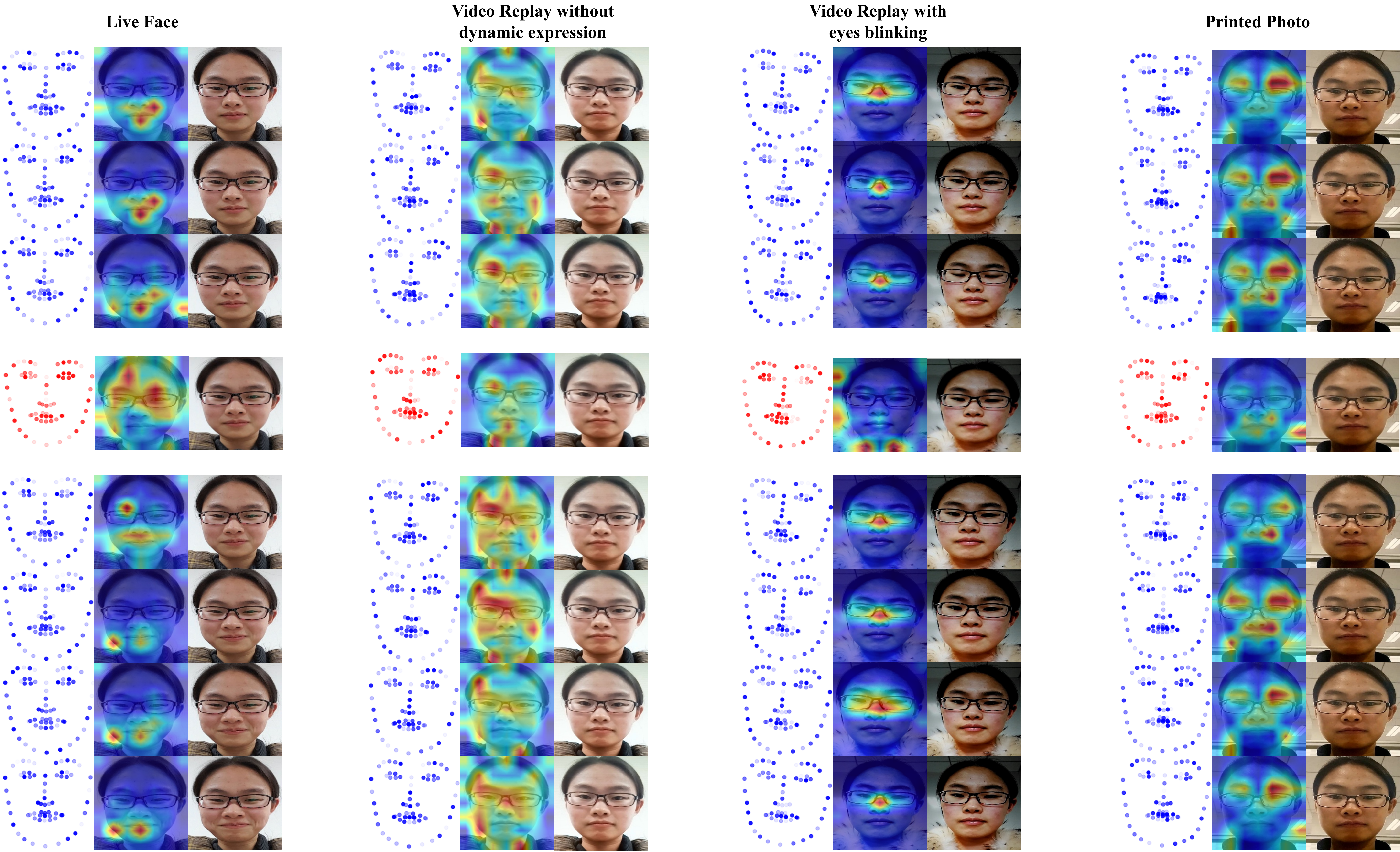}
    \caption{Visualization examples. Turning to live face videos, spatial attention highlights the entire facial area, while temporal attention specifically detects mouth movements. For video replay attacks lacking dynamic expressions, spatial attention and temporal attention both focus on entire face area. In the case of video replay attacks involving eye blinking, spatial attention is locked to abnormal photometric clue, while temporal attention is aware of the eye blinking. Through information fusion, the system should ultimately be able to distinguish such instances as spoofed. Turning to printed photos, spatial attention concentrates on the part of facial area and backgrounds lighting abnormal, however, temporal attention searches for the eyes and mouth of each photo to detect movement.}
    \label{fig: Visualizing}
\end{figure*}

\section{Visualization and Analysis}
\label{subsec: visualization and analysis}

\noindent \textbf{Feature Distribution.} 
The distribution of fusion head token is visualized by t-SNE \cite{van2008visualizing}, as shown in Fig. \ref{fig: TSNE}. A reminder is that we take the head tokens for each frame without implementing pooling operations on them. We visualize the feature distributions under the O\&C\&I to M and O\&C\&I to 3DM protocols. From the figures, we can observe a good alignment between the target domains and the source domains in both, and the distributions also have some similarities. This demonstrates dynamic and static combinations of anti-spoofing representations between different domains and attack types have high consistency and also provide a powerful explanation for the excellent generalization ability of the model.

\noindent \textbf{Spatiotemporal Attention Visualization.}
We sample 8 frames of live faces and 3D masks to illustrate the attention map. In Fig. \ref{fig: Explainability}, the spatial attention reveals which regions of the $t$-th frame are highlighted, while the temporal attention shows which regions of other frames (except for $t$-th frame) receive more focus. In spatial attention, we observe that the red response area corresponds to the several facial area and background, responsible for capturing photometric information, potentially related to materials, textures, or lighting. However, temporal attention is more concentrated on motion areas, such as the mouth's motion in live videos, or the awareness of the absence of variations of facial expressions in spoof videos. We utilize the Transformer Explainability \cite{chefer2021transformer} to present the visualization and provide a brief description of the process of generating attention maps. For each attention layer, we can obtain the attention map \(A^{(b)}\) and its gradients \(\nabla A^{(b)}\), and relevance \(R^{n_b}\), where \(n_b\) is the layer that corresponds to the \(\mathrm{Softmax}\) operation in Eq. \ref{eqn: attention} of transformer block \(b\), and \(R^{n_b}\) is the layer's relevance. Following the propagation rule of relevance and gradient illustrated on the upper right of Fig. \ref{fig: Explainability}, we can conclude the final relevancy maps in Eq. \ref{eqn: relevance},\ref{eqn: rollout}: 
\begin{equation}
\label{eqn: attention}
A^{(b)} = \mathrm{Softmax}(Q^{(b)}K^{(b)T}/\sqrt{d_k}),
\end{equation}
\begin{equation}
\label{eqn: attention output}
O^{(b)} = A^{(b)}V^{(b)},
\end{equation}
\begin{equation}
\label{eqn: relevance}
\bar{A}^{(b)} = I+\mathbb{E}_h(\nabla A^{(b)} \odot R^{n_b})^+,
\end{equation}
\begin{equation}
\label{eqn: rollout}
C = \bar{A}^{(1)} \cdot \bar{A}^{(2)}\cdot \cdot \cdot \bar{A}^{(B)},
\end{equation}
where \(\odot\) is the Hadamard product, and \(\mathbb{E}_h\) is the mean across the "heads" dimension. \(C\) is the final output for visualization.

For the spatial attention's relevancy map of $t$-th frame, its shape is \([L+1,L+1]\). We take the attention weight corresponding to the first row except for the class token itself, with a length of \(L\), where \(L\) is the number of patches, and reshape it into 2D. By upsampling, we can obtain the spatial attention map. While the shape of temporal attention's relevancy map is \([T\times (L+1),T\times (L+1)]\), where \(T\) is the number of frames. We take the time segment from \(t-\tau\) to \(t+\tau\) as an example, because masking of the same frame patch by our Kronecker temporal attention, we only need to take attention weights in different timestamps. Two weight sequences with a length of \(L\) correspond to the temporal attention map of frame \(t\) regarding \(t-\tau\) and \(t+\tau\). We use the same method to obtain graph attention maps. In terms of time dimension, the model is more interested in the mouth, eyes, and eyebrow bones because these areas have a higher probability of relative movement. In spatial attention, the areas of interest are relatively scattered, dedicated to perceiving the topology of facial landmarks. 

More spatiotemporal attention visualization maps are shown in Fig. \ref{fig: Visualizing}, including real samples, photo printing attacks, video replay without dynamic expression and video replay with eyes blinking attacks. Regarding live face videos, spatial attention highlights the facial area (forehead, cheeks), aggregating photometric information. While temporal attention particularly detects mouth movements (smiling). For video replay attacks lacking dynamic expressions, spatial attention and temporal attention both focus on entire face area. In the case of video replay attacks involving eye blinking, spatial attention focuses on abnormal photometric cues caused by electronic device screens, while temporal attention is aware of the eye blinking. Through spatiotemporal information fusion, the model is ultimately able to distinguish whether such a instance is a spoof sample. For printing attacks, spatial attention locates facial areas and backgrounds with abnormal lighting. At the same time, in the temporal dimension, temporal attention searches for the eyes and mouth of each photo but cannot detect any tiny movements.

\section{Conclusion}
\label{sec:conclusion}
\noindent This study introduces a pioneering architecture Graph Guided Video Vision Transformer (G\(^2\)V\(^2\)former) to capture discriminative spatio-temporal spoofing shreds of evidence for the FAS task by concurrently leveraging facial images and landmarks. We factorize the joint attention into spatial and temporal attention, utilizing photometric consistency loss to guide the spatial attention. We propose a novel Kronecker temporal attention with a wider receptive field, and then utilize the motion trajectory contained in facial landmarks to guide the pixel motion capture. Through extensive experiments and ablation studies, we demonstrate the effectiveness of G\(^2\)V\(^2\)former which achieves extraordinary generalization even in more challenging scenarios.

\bibliographystyle{IEEEtran}
\bibliography{main}

\begin{thebibliography}{10}
\providecommand{\url}[1]{#1}
\csname url@samestyle\endcsname
\providecommand{\newblock}{\relax}
\providecommand{\bibinfo}[2]{#2}
\providecommand{\BIBentrySTDinterwordspacing}{\spaceskip=0pt\relax}
\providecommand{\BIBentryALTinterwordstretchfactor}{4}
\providecommand{\BIBentryALTinterwordspacing}{\spaceskip=\fontdimen2\font plus
\BIBentryALTinterwordstretchfactor\fontdimen3\font minus \fontdimen4\font\relax}
\providecommand{\BIBforeignlanguage}[2]{{%
\expandafter\ifx\csname l@#1\endcsname\relax
\typeout{** WARNING: IEEEtran.bst: No hyphenation pattern has been}%
\typeout{** loaded for the language `#1'. Using the pattern for}%
\typeout{** the default language instead.}%
\else
\language=\csname l@#1\endcsname
\fi
#2}}
\providecommand{\BIBdecl}{\relax}
\BIBdecl

\bibitem{boulkenafet2015face}
Z.~Boulkenafet, J.~Komulainen, and A.~Hadid, ``Face anti-spoofing based on color texture analysis,'' in \emph{2015 IEEE international conference on image processing (ICIP)}.\hskip 1em plus 0.5em minus 0.4em\relax IEEE, 2015, pp. 2636--2640.

\bibitem{patel2016secure}
K.~Patel, H.~Han, and A.~K. Jain, ``Secure face unlock: Spoof detection on smartphones,'' \emph{IEEE transactions on information forensics and security}, vol.~11, no.~10, pp. 2268--2283, 2016.

\bibitem{yang2013face}
J.~Yang, Z.~Lei, S.~Liao, and S.~Z. Li, ``Face liveness detection with component dependent descriptor,'' in \emph{2013 International Conference on Biometrics (ICB)}.\hskip 1em plus 0.5em minus 0.4em\relax IEEE, 2013, pp. 1--6.

\bibitem{li2016original}
L.~Li, X.~Feng, Z.~Boulkenafet, Z.~Xia, M.~Li, and A.~Hadid, ``An original face anti-spoofing approach using partial convolutional neural network,'' in \emph{2016 Sixth International Conference on Image Processing Theory, Tools and Applications (IPTA)}.\hskip 1em plus 0.5em minus 0.4em\relax IEEE, 2016, pp. 1--6.

\bibitem{yu2020searching}
Z.~Yu, C.~Zhao, Z.~Wang, Y.~Qin, Z.~Su, X.~Li, F.~Zhou, and G.~Zhao, ``Searching central difference convolutional networks for face anti-spoofing,'' in \emph{Proceedings of the IEEE/CVF Conference on Computer Vision and Pattern Recognition}, 2020, pp. 5295--5305.

\bibitem{shao2019multi}
R.~Shao, X.~Lan, J.~Li, and P.~C. Yuen, ``Multi-adversarial discriminative deep domain generalization for face presentation attack detection,'' in \emph{Proceedings of the IEEE/CVF conference on computer vision and pattern recognition}, 2019, pp. 10\,023--10\,031.

\bibitem{jia2020single}
Y.~Jia, J.~Zhang, S.~Shan, and X.~Chen, ``Single-side domain generalization for face anti-spoofing,'' in \emph{Proceedings of the IEEE/CVF Conference on Computer Vision and Pattern Recognition}, 2020, pp. 8484--8493.

\bibitem{chen2021generalizable}
Z.~Chen, T.~Yao, K.~Sheng, S.~Ding, Y.~Tai, J.~Li, F.~Huang, and X.~Jin, ``Generalizable representation learning for mixture domain face anti-spoofing,'' in \emph{Proceedings of the AAAI Conference on Artificial Intelligence}, vol.~35, no.~2, 2021, pp. 1132--1139.

\bibitem{liu2021adaptive}
S.~Liu, K.-Y. Zhang, T.~Yao, M.~Bi, S.~Ding, J.~Li, F.~Huang, and L.~Ma, ``Adaptive normalized representation learning for generalizable face anti-spoofing,'' in \emph{Proceedings of the 29th ACM international conference on multimedia}, 2021, pp. 1469--1477.

\bibitem{wang2022domain}
Z.~Wang, Z.~Wang, Z.~Yu, W.~Deng, J.~Li, T.~Gao, and Z.~Wang, ``Domain generalization via shuffled style assembly for face anti-spoofing,'' in \emph{Proceedings of the IEEE/CVF Conference on Computer Vision and Pattern Recognition}, 2022, pp. 4123--4133.

\bibitem{zhou2023instance}
Q.~Zhou, K.-Y. Zhang, T.~Yao, X.~Lu, R.~Yi, S.~Ding, and L.~Ma, ``Instance-aware domain generalization for face anti-spoofing,'' in \emph{Proceedings of the IEEE/CVF Conference on Computer Vision and Pattern Recognition}, 2023, pp. 20\,453--20\,463.

\bibitem{sun2023rethinking}
Y.~Sun, Y.~Liu, X.~Liu, Y.~Li, and W.-S. Chu, ``Rethinking domain generalization for face anti-spoofing: Separability and alignment,'' in \emph{Proceedings of the IEEE/CVF Conference on Computer Vision and Pattern Recognition}, 2023, pp. 24\,563--24\,574.

\bibitem{yang2024generalized}
J.~Yang, Z.~Yu, X.~Ni, J.~He, and H.~Li, ``Generalized face anti-spoofing via finer domain partition and disentangling liveness-irrelevant factors,'' \emph{European Conference on Artificial Intelligence (ECAI)}, 2024.

\bibitem{wang2020unsupervised}
G.~Wang, H.~Han, S.~Shan, and X.~Chen, ``Unsupervised adversarial domain adaptation for cross-domain face presentation attack detection,'' \emph{IEEE Transactions on Information Forensics and Security}, vol.~16, pp. 56--69, 2020.

\bibitem{wang2021self}
J.~Wang, J.~Zhang, Y.~Bian, Y.~Cai, C.~Wang, and S.~Pu, ``Self-domain adaptation for face anti-spoofing,'' in \emph{Proceedings of the AAAI Conference on Artificial Intelligence}, vol.~35, no.~4, 2021, pp. 2746--2754.

\bibitem{shao2020regularized}
R.~Shao, X.~Lan, and P.~C. Yuen, ``Regularized fine-grained meta face anti-spoofing,'' in \emph{Proceedings of the AAAI Conference on Artificial Intelligence}, vol.~34, no.~07, 2020, pp. 11\,974--11\,981.

\bibitem{liu2018learning}
Y.~Liu, A.~Jourabloo, and X.~Liu, ``Learning deep models for face anti-spoofing: Binary or auxiliary supervision,'' in \emph{Proceedings of the IEEE conference on computer vision and pattern recognition}, 2018, pp. 389--398.

\bibitem{saha2020domain}
S.~Saha, W.~Xu, M.~Kanakis, S.~Georgoulis, Y.~Chen, D.~P. Paudel, and L.~Van~Gool, ``Domain agnostic feature learning for image and video based face anti-spoofing,'' in \emph{Proceedings of the IEEE/CVF Conference on Computer Vision and Pattern Recognition Workshops}, 2020, pp. 802--803.

\bibitem{wang2020deep}
Z.~Wang, Z.~Yu, C.~Zhao, X.~Zhu, Y.~Qin, Q.~Zhou, F.~Zhou, and Z.~Lei, ``Deep spatial gradient and temporal depth learning for face anti-spoofing,'' in \emph{Proceedings of the IEEE/CVF Conference on Computer Vision and Pattern Recognition}, 2020, pp. 5042--5051.

\bibitem{yang2019face}
X.~Yang, W.~Luo, L.~Bao, Y.~Gao, D.~Gong, S.~Zheng, Z.~Li, and W.~Liu, ``Face anti-spoofing: Model matters, so does data,'' in \emph{Proceedings of the IEEE/CVF Conference on Computer Vision and Pattern Recognition}, 2019, pp. 3507--3516.

\bibitem{wang2021multi}
Z.~Wang, Y.~Xu, L.~Wu, H.~Han, Y.~Ma, and G.~Ma, ``Multi-perspective features learning for face anti-spoofing,'' in \emph{Proceedings of the IEEE/CVF International Conference on Computer Vision}, 2021, pp. 4116--4122.

\bibitem{xu2021exploiting}
Y.~Xu, Z.~Wang, H.~Han, L.~Wu, and Y.~Liu, ``Exploiting non-uniform inherent cues to improve presentation attack detection,'' in \emph{2021 IEEE International Joint Conference on Biometrics (IJCB)}.\hskip 1em plus 0.5em minus 0.4em\relax IEEE, 2021, pp. 1--8.

\bibitem{ming2022vitranspad}
Z.~Ming, Z.~Yu, M.~Al-Ghadi, M.~Visani, M.~M. Luqman, and J.-C. Burie, ``Vitranspad: video transformer using convolution and self-attention for face presentation attack detection,'' in \emph{2022 IEEE International Conference on Image Processing (ICIP)}.\hskip 1em plus 0.5em minus 0.4em\relax IEEE, 2022, pp. 4248--4252.

\bibitem{khan2021video}
S.~A. Khan and H.~Dai, ``Video transformer for deepfake detection with incremental learning,'' in \emph{Proceedings of the 29th ACM International Conference on Multimedia}, 2021, pp. 1821--1828.

\bibitem{wang2022learning}
Z.~Wang, Q.~Wang, W.~Deng, and G.~Guo, ``Learning multi-granularity temporal characteristics for face anti-spoofing,'' \emph{IEEE Transactions on Information Forensics and Security}, vol.~17, pp. 1254--1269, 2022.

\bibitem{chang2023closer}
C.-J. Chang, Y.-C. Lee, S.-H. Yao, M.-H. Chen, C.-Y. Wang, S.-H. Lai, and T.~P.-C. Chen, ``A closer look at geometric temporal dynamics for face anti-spoofing,'' in \emph{Proceedings of the IEEE/CVF Conference on Computer Vision and Pattern Recognition Biometrics Workshop}, 2023, pp. 1081--1091.

\bibitem{yan2018spatial}
S.~Yan, Y.~Xiong, and D.~Lin, ``Spatial temporal graph convolutional networks for skeleton-based action recognition,'' in \emph{Proceedings of the AAAI conference on artificial intelligence}, vol.~32, no.~1, 2018.

\bibitem{li2016generalized}
X.~Li, J.~Komulainen, G.~Zhao, P.-C. Yuen, and M.~Pietik{\"a}inen, ``Generalized face anti-spoofing by detecting pulse from face videos,'' in \emph{2016 23rd International Conference on Pattern Recognition (ICPR)}.\hskip 1em plus 0.5em minus 0.4em\relax IEEE, 2016, pp. 4244--4249.

\bibitem{liu2018remote}
S.-Q. Liu, X.~Lan, and P.~C. Yuen, ``Remote photoplethysmography correspondence feature for 3d mask face presentation attack detection,'' in \emph{Proceedings of the European Conference on Computer Vision (ECCV)}, 2018, pp. 558--573.

\bibitem{yu2021transrppg}
Z.~Yu, X.~Li, P.~Wang, and G.~Zhao, ``Transrppg: Remote photoplethysmography transformer for 3d mask face presentation attack detection,'' \emph{IEEE Signal Processing Letters}, vol.~28, pp. 1290--1294, 2021.

\bibitem{patel2016cross}
K.~Patel, H.~Han, and A.~K. Jain, ``Cross-database face antispoofing with robust feature representation,'' in \emph{Biometric Recognition: 11th Chinese Conference, CCBR 2016, Chengdu, China, October 14-16, 2016, Proceedings 11}.\hskip 1em plus 0.5em minus 0.4em\relax Springer, 2016, pp. 611--619.

\bibitem{kollreider2007real}
K.~Kollreider, H.~Fronthaler, M.~I. Faraj, and J.~Bigun, ``Real-time face detection and motion analysis with application in “liveness” assessment,'' \emph{IEEE Transactions on Information Forensics and Security}, vol.~2, no.~3, pp. 548--558, 2007.

\bibitem{shao2017deep}
R.~Shao, X.~Lan, and P.~C. Yuen, ``Deep convolutional dynamic texture learning with adaptive channel-discriminability for 3d mask face anti-spoofing,'' in \emph{2017 IEEE International Joint Conference on Biometrics (IJCB)}.\hskip 1em plus 0.5em minus 0.4em\relax IEEE, 2017, pp. 748--755.

\bibitem{dosovitskiy2020image}
A.~Dosovitskiy, L.~Beyer, A.~Kolesnikov, D.~Weissenborn, X.~Zhai, T.~Unterthiner, M.~Dehghani, M.~Minderer, G.~Heigold, S.~Gelly \emph{et~al.}, ``An image is worth 16x16 words: Transformers for image recognition at scale,'' \emph{arXiv preprint arXiv:2010.11929}, 2020.

\bibitem{bertasius2021space}
G.~Bertasius, H.~Wang, and L.~Torresani, ``Is space-time attention all you need for video understanding?'' in \emph{ICML}, vol.~2, no.~3, 2021, p.~4.

\bibitem{arnab2021vivit}
A.~Arnab, M.~Dehghani, G.~Heigold, C.~Sun, M.~Lu{\v{c}}i{\'c}, and C.~Schmid, ``Vivit: A video vision transformer,'' in \emph{Proceedings of the IEEE/CVF international conference on computer vision}, 2021, pp. 6836--6846.

\bibitem{neimark2021video}
D.~Neimark, O.~Bar, M.~Zohar, and D.~Asselmann, ``Video transformer network,'' in \emph{Proceedings of the IEEE/CVF international conference on computer vision}, 2021, pp. 3163--3172.

\bibitem{guo2021ssan}
X.~Guo, X.~Guo, and Y.~Lu, ``Ssan: Separable self-attention network for video representation learning,'' in \emph{Proceedings of the IEEE/CVF conference on computer vision and pattern recognition}, 2021, pp. 12\,618--12\,627.

\bibitem{liu2022video}
Z.~Liu, J.~Ning, Y.~Cao, Y.~Wei, Z.~Zhang, S.~Lin, and H.~Hu, ``Video swin transformer,'' in \emph{Proceedings of the IEEE/CVF conference on computer vision and pattern recognition}, 2022, pp. 3202--3211.

\bibitem{yang2025kronecker}
J.~Yang, Z.~Yu, X.~Ni, J.~He, and H.~Li, ``Kronecker mask and interpretive prompts are language-action video learners,'' \emph{arXiv preprint arXiv:2502.03549}, 2025.

\bibitem{he2022masked}
K.~He, X.~Chen, S.~Xie, Y.~Li, P.~Doll{\'a}r, and R.~Girshick, ``Masked autoencoders are scalable vision learners,'' in \emph{Proceedings of the IEEE/CVF conference on computer vision and pattern recognition}, 2022, pp. 16\,000--16\,009.

\bibitem{feichtenhofer2022masked}
C.~Feichtenhofer, Y.~Li, K.~He \emph{et~al.}, ``Masked autoencoders as spatiotemporal learners,'' \emph{Advances in neural information processing systems}, vol.~35, pp. 35\,946--35\,958, 2022.

\bibitem{tong2022videomae}
Z.~Tong, Y.~Song, J.~Wang, and L.~Wang, ``Videomae: Masked autoencoders are data-efficient learners for self-supervised video pre-training,'' \emph{Advances in neural information processing systems}, vol.~35, pp. 10\,078--10\,093, 2022.

\bibitem{kipf2016semi}
T.~N. Kipf and M.~Welling, ``Semi-supervised classification with graph convolutional networks,'' \emph{arXiv preprint arXiv:1609.02907}, 2016.

\bibitem{chen2021channel}
Y.~Chen, Z.~Zhang, C.~Yuan, B.~Li, Y.~Deng, and W.~Hu, ``Channel-wise topology refinement graph convolution for skeleton-based action recognition,'' in \emph{Proceedings of the IEEE/CVF international conference on computer vision}, 2021, pp. 13\,359--13\,368.

\bibitem{liu2020disentangling}
Y.~Liu, J.~Stehouwer, and X.~Liu, ``On disentangling spoof trace for generic face anti-spoofing,'' in \emph{Computer Vision--ECCV 2020: 16th European Conference, Glasgow, UK, August 23--28, 2020, Proceedings, Part XVIII 16}.\hskip 1em plus 0.5em minus 0.4em\relax Springer, 2020, pp. 406--422.

\bibitem{shi2019two}
L.~Shi, Y.~Zhang, J.~Cheng, and H.~Lu, ``Two-stream adaptive graph convolutional networks for skeleton-based action recognition,'' in \emph{Proceedings of the IEEE/CVF conference on computer vision and pattern recognition}, 2019, pp. 12\,026--12\,035.

\bibitem{ying2021transformers}
C.~Ying, T.~Cai, S.~Luo, S.~Zheng, G.~Ke, D.~He, Y.~Shen, and T.-Y. Liu, ``Do transformers really perform badly for graph representation?'' \emph{Advances in Neural Information Processing Systems}, vol.~34, pp. 28\,877--28\,888, 2021.

\bibitem{mildenhall2021nerf}
B.~Mildenhall, P.~P. Srinivasan, M.~Tancik, J.~T. Barron, R.~Ramamoorthi, and R.~Ng, ``Nerf: Representing scenes as neural radiance fields for view synthesis,'' \emph{Communications of the ACM}, vol.~65, no.~1, pp. 99--106, 2021.

\bibitem{liu2021swin}
Z.~Liu, Y.~Lin, Y.~Cao, H.~Hu, Y.~Wei, Z.~Zhang, S.~Lin, and B.~Guo, ``Swin transformer: Hierarchical vision transformer using shifted windows,'' in \emph{Proceedings of the IEEE/CVF international conference on computer vision}, 2021, pp. 10\,012--10\,022.

\bibitem{boulkenafet2017oulu}
Z.~Boulkenafet, J.~Komulainen, L.~Li, X.~Feng, and A.~Hadid, ``Oulu-npu: A mobile face presentation attack database with real-world variations,'' in \emph{2017 12th IEEE international conference on automatic face \& gesture recognition (FG 2017)}.\hskip 1em plus 0.5em minus 0.4em\relax IEEE, 2017, pp. 612--618.

\bibitem{zhang2012face}
Z.~Zhang, J.~Yan, S.~Liu, Z.~Lei, D.~Yi, and S.~Z. Li, ``A face antispoofing database with diverse attacks,'' in \emph{2012 5th IAPR international conference on Biometrics (ICB)}.\hskip 1em plus 0.5em minus 0.4em\relax IEEE, 2012, pp. 26--31.

\bibitem{chingovska2012effectiveness}
I.~Chingovska, A.~Anjos, and S.~Marcel, ``On the effectiveness of local binary patterns in face anti-spoofing,'' in \emph{2012 BIOSIG-proceedings of the international conference of biometrics special interest group (BIOSIG)}.\hskip 1em plus 0.5em minus 0.4em\relax IEEE, 2012, pp. 1--7.

\bibitem{wen2015face}
D.~Wen, H.~Han, and A.~K. Jain, ``Face spoof detection with image distortion analysis,'' \emph{IEEE Transactions on Information Forensics and Security}, vol.~10, no.~4, pp. 746--761, 2015.

\bibitem{zhang2019dataset}
S.~Zhang, X.~Wang, A.~Liu, C.~Zhao, J.~Wan, S.~Escalera, H.~Shi, Z.~Wang, and S.~Z. Li, ``A dataset and benchmark for large-scale multi-modal face anti-spoofing,'' in \emph{Proceedings of the IEEE/CVF Conference on Computer Vision and Pattern Recognition}, 2019, pp. 919--928.

\bibitem{zhang2020casia}
S.~Zhang, A.~Liu, J.~Wan, Y.~Liang, G.~Guo, S.~Escalera, H.~J. Escalante, and S.~Z. Li, ``Casia-surf: A large-scale multi-modal benchmark for face anti-spoofing,'' \emph{IEEE Transactions on Biometrics, Behavior, and Identity Science}, vol.~2, no.~2, pp. 182--193, 2020.

\bibitem{liu2021casia}
A.~Liu, Z.~Tan, J.~Wan, S.~Escalera, G.~Guo, and S.~Z. Li, ``Casia-surf cefa: A benchmark for multi-modal cross-ethnicity face anti-spoofing,'' in \emph{Proceedings of the IEEE/CVF winter conference on applications of computer vision}, 2021, pp. 1179--1187.

\bibitem{liu2021cross}
A.~Liu, X.~Li, J.~Wan, Y.~Liang, S.~Escalera, H.~J. Escalante, M.~Madadi, Y.~Jin, Z.~Wu, X.~Yu \emph{et~al.}, ``Cross-ethnicity face anti-spoofing recognition challenge: A review,'' \emph{IET Biometrics}, vol.~10, no.~1, pp. 24--43, 2021.

\bibitem{george2019biometric}
A.~George, Z.~Mostaani, D.~Geissenbuhler, O.~Nikisins, A.~Anjos, and S.~Marcel, ``Biometric face presentation attack detection with multi-channel convolutional neural network,'' \emph{IEEE transactions on information forensics and security}, vol.~15, pp. 42--55, 2019.

\bibitem{erdogmus2013spoofing}
N.~Erdogmus and S.~Marcel, ``Spoofing in 2d face recognition with 3d masks and anti-spoofing with kinect,'' in \emph{2013 IEEE sixth international conference on biometrics: theory, applications and systems (BTAS)}.\hskip 1em plus 0.5em minus 0.4em\relax IEEE, 2013, pp. 1--6.

\bibitem{liu20163d}
S.~Liu, P.~C. Yuen, S.~Zhang, and G.~Zhao, ``3d mask face anti-spoofing with remote photoplethysmography,'' in \emph{Computer Vision--ECCV 2016: 14th European Conference, Amsterdam, The Netherlands, October 11--14, 2016, Proceedings, Part VII 14}.\hskip 1em plus 0.5em minus 0.4em\relax Springer, 2016, pp. 85--100.

\bibitem{zhang2016joint}
K.~Zhang, Z.~Zhang, Z.~Li, and Y.~Qiao, ``Joint face detection and alignment using multitask cascaded convolutional networks,'' \emph{IEEE signal processing letters}, vol.~23, no.~10, pp. 1499--1503, 2016.

\bibitem{guo2020towards}
J.~Guo, X.~Zhu, Y.~Yang, F.~Yang, Z.~Lei, and S.~Z. Li, ``Towards fast, accurate and stable 3d dense face alignment,'' in \emph{European Conference on Computer Vision}.\hskip 1em plus 0.5em minus 0.4em\relax Springer, 2020, pp. 152--168.

\bibitem{kingma2014adam}
D.~P. Kingma and J.~Ba, ``Adam: A method for stochastic optimization,'' \emph{arXiv preprint arXiv:1412.6980}, 2014.

\bibitem{loshchilov2016sgdr}
I.~Loshchilov and F.~Hutter, ``Sgdr: Stochastic gradient descent with warm restarts,'' \emph{arXiv preprint arXiv:1608.03983}, 2016.

\bibitem{li2018domain}
H.~Li, S.~J. Pan, S.~Wang, and A.~C. Kot, ``Domain generalization with adversarial feature learning,'' in \emph{Proceedings of the IEEE conference on computer vision and pattern recognition}, 2018, pp. 5400--5409.

\bibitem{yu2020fas}
Z.~Yu, J.~Wan, Y.~Qin, X.~Li, S.~Z. Li, and G.~Zhao, ``Nas-fas: Static-dynamic central difference network search for face anti-spoofing,'' \emph{IEEE transactions on pattern analysis and machine intelligence}, vol.~43, no.~9, pp. 3005--3023, 2020.

\bibitem{liu2021dual}
S.~Liu, K.-Y. Zhang, T.~Yao, K.~Sheng, S.~Ding, Y.~Tai, J.~Li, Y.~Xie, and L.~Ma, ``Dual reweighting domain generalization for face presentation attack detection,'' \emph{arXiv preprint arXiv:2106.16128}, 2021.

\bibitem{wang2022patchnet}
C.-Y. Wang, Y.-D. Lu, S.-T. Yang, and S.-H. Lai, ``Patchnet: A simple face anti-spoofing framework via fine-grained patch recognition,'' in \emph{Proceedings of the IEEE/CVF Conference on Computer Vision and Pattern Recognition}, 2022, pp. 20\,281--20\,290.

\bibitem{liu2023towards}
Y.~Liu, Y.~Chen, M.~Gou, C.-T. Huang, Y.~Wang, W.~Dai, and H.~Xiong, ``Towards unsupervised domain generalization for face anti-spoofing,'' in \emph{Proceedings of the IEEE/CVF International Conference on Computer Vision}, 2023, pp. 20\,654--20\,664.

\bibitem{zhou2024test}
Q.~Zhou, K.-Y. Zhang, T.~Yao, X.~Lu, S.~Ding, and L.~Ma, ``Test-time domain generalization for face anti-spoofing,'' in \emph{Proceedings of the IEEE/CVF Conference on Computer Vision and Pattern Recognition}, 2024, pp. 175--187.

\bibitem{le2024gradient}
B.~M. Le and S.~S. Woo, ``Gradient alignment for cross-domain face anti-spoofing,'' in \emph{Proceedings of the IEEE/CVF Conference on Computer Vision and Pattern Recognition}, 2024, pp. 188--199.

\bibitem{huang2022adaptive}
H.-P. Huang, D.~Sun, Y.~Liu, W.-S. Chu, T.~Xiao, J.~Yuan, H.~Adam, and M.-H. Yang, ``Adaptive transformers for robust few-shot cross-domain face anti-spoofing,'' in \emph{European conference on computer vision}.\hskip 1em plus 0.5em minus 0.4em\relax Springer, 2022, pp. 37--54.

\bibitem{he2016deep}
K.~He, X.~Zhang, S.~Ren, and J.~Sun, ``Deep residual learning for image recognition,'' in \emph{Proceedings of the IEEE conference on computer vision and pattern recognition}, 2016, pp. 770--778.

\bibitem{chefer2021transformer}
H.~Chefer, S.~Gur, and L.~Wolf, ``Transformer interpretability beyond attention visualization,'' in \emph{Proceedings of the IEEE/CVF conference on computer vision and pattern recognition}, 2021, pp. 782--791.

\bibitem{van2008visualizing}
L.~Van~der Maaten and G.~Hinton, ``Visualizing data using t-sne.'' \emph{Journal of machine learning research}, vol.~9, no.~11, 2008.

\end{thebibliography}
\end{document}